\documentclass{clv2-eng}
\usepackage{url}
\usepackage{latexsym}

\usepackage[english]{babel}

\usepackage{covington}
\usepackage{graphicx}
\usepackage{subfigure}
\usepackage{tabularx}
\usepackage{booktabs}
\usepackage[table]{xcolor}
\definecolor{LighterGray}{gray}{0.92}

\usepackage{tikz}
\usepackage{tikz-qtree}
\usetikzlibrary{arrows}

\usepackage{ragged2e}

\usepackage{amsmath}
\DeclareMathOperator*{\argmin}{arg\,min}

\runningtitle{Distributed Representations of Lexical Sets and Prototypes}

\runningauthor{Ponti et al.}

\title{Distributed Representations of Lexical Sets and Prototypes in Causal Alternation Verbs}

\begin{document}

\author{Edoardo Maria Ponti\thanks{\texttt{ep490@cam.ac.uk}}}
\affil{Universit\'a di Cambridge}
\author{Elisabetta Jezek\thanks{\texttt{jezek@unipv.it}}}
\affil{Universit\`a degli Studi di Pavia}
\author{Bernardo Magnini\thanks{\texttt{magnini@fbk.eu}}}
\affil{Fondazione Bruno Kessler}

\maketitle

\begin{abstract}

\noindent Lexical sets contain the words filling an argument slot of a verb, and are in part determined by selectional preferences. The purpose of this paper is to unravel the properties of lexical sets through distributional semantics. We investigate 1) whether lexical set behave as prototypical categories with a centre and a periphery; 2) whether they are polymorphic, i.e.\ composed by sub-categories; 3) whether the distance between lexical sets of different arguments is explanatory of verb properties. In particular, our case study are lexical sets of causative-inchoative verbs in Italian. Having studied several vector models, we find that 1) based on spatial distance from the centroid, object fillers are scattered uniformly across the category, whereas intransitive subject fillers lie on its edge; 2) a correlation exists between the amount of verb senses and that of clusters discovered automatically, especially for intransitive subjects; 3) the distance between the centroids of object and intransitive subject is correlated with other properties of verbs, such as their cross-lingual tendency to appear in the intransitive pattern rather than transitive one. This paper is noncommittal with respect to the hypothesis that this connection is underpinned by a semantic reason, namely the spontaneity of the event denoted by the verb. 
\end{abstract}

\section{Introduction}
\label{sec:intro}
The arguments of a verb are the ``slots'' that have to be filled to satisfy its valency (subject, object, etc.). Hence, verbs display so-called selectional restrictions over the possible fillers occupying these slots \cite{seaghdha2014probabilistic}, which play a major role in determining the verb meaning. Moreover, the selection of the fillers happens in accordance with the different senses of a verb. Lists of the possible fillers occurring with a certain pattern of a verb can be collected in a corpus-driven fashion: these lists are named ``lexical sets'' \cite{hanks1996}.

Several approaches in Computational Linguistics managed to inspect lexical sets and their patterns of variation \cite{montemagni1995,mckoon2000}, as well as to use them as features for verb classification \cite{mccarthy2000using,joanis2008general}. On the other hand, selectional preferences were employed for many tasks in Natural Language Processing, including Word Sense Disambiguation \cite[\textit{inter alia}]{resnik1997selectional,mccarthy2003disambiguating}, Metaphor Processing \cite{shutova2013statistical},
Information Extraction \cite{pantel2007isp}, Discourse Relation Classification \cite{ponti2017event}, Dependency Parsing \cite{mirroshandel2016integrating}, and Semantic Role labeling \cite{gildea2002automatic,zapirain2013selectional}. 

We aim at establishing a new method of analysis for lexical sets. In particular, we address the following key questions: are lexical sets Aristotelian categories (yes-no membership) or prototypical categories (graded membership)? If the latter holds true, are the fillers arranged homogeneously or do they cluster around some sub-categories? Finally, is the relation between lexical sets from different patterns of a same verb informative about this verb's meaning? We address these questions under a distributional semantics perspective. In fact, distributional semantics provides a mapping between each filler and a vector lying in a continuous multi-dimensional space. By virtue of this, lexical sets can be treated as continuous categories, where members can be either central or peripheral. Moreover, this allows to quantify the distance between vectors with spatial measures.

In order to test the relevance for linguistic theory of this approach, we focus on a case study, namely verbs undergoing the causative-inchoative alternation. Solving the above-mentioned issues may help clarifying some of the vexed questions about this class of verbs. These show both transitive and intransitive patterns: the object of the former and the subject of the latter play the same semantic role of patient. Based on the cross-lingual frequency of each of these patterns and the direction of morphological derivation, it is possible to sort these verbs onto a scale \cite{haspelmath1993,samardzic2012meaning,haspelmath2014coding}, which can be possibly interpreted semantically as the ``spontaneity'' of the corresponding event (see \S\ \ref{ssec:cauinc}). We investigate the existence of any asymmetry between the lexical sets of the transitive object and the intransitive subject, and if so the connection of this asymmetry with the spontaneity scale. 

The structure of the paper is as follows. In \S\ \ref{sec:definitions}, we define the core notions of this study, including lexical sets, causative-inchoative verbs, and distributional semantics. \S\ \ref{sec:data} presents the method and the data, whereas \S\ \ref{sec:experiments} reports the results of the experiments. Finally, \S\ \ref{sec:discussion} draws the conclusions and \S\ \ref{sec:conclusion} proposes possible future lines of research.

\section{Definitions and Previous Work}
\label{sec:definitions}
In this section, we describe in detail the notions underlying the subject matter of the research (lexical sets, \S\ \ref{ssec:lexset}), the case study (causative-inchoative verbs, \S\ \ref{ssec:cauinc}), and the method (distributional semantics, \S\ \ref{ssec:we}). At the same time, we present the previous work concerning each of these aspects.

\subsection{Lexical Sets}
\label{ssec:lexset}

A lexical set can be defined as the set of words that occupy a specific argument position within a single verb sense, such as \{{\it gun}, {\it bullet}, {\it shot}, {\it projectile}, {\it rifle}...\} for the sense `to shoot' of {\it to fire}, or \{{\it employer}, {\it teacher}, {\it attorney}, {\it manager}...\} for its sense `to dismiss'. The notion of lexical set was firstly introduced by \namecite{hanks1996}. Its purpose is explaining how the semantics of verbs is affected by the patterns of complements they are found with. Hanks' approach is justified by the pervasiveness of patterns in corpora: these patterns are instantiated by specific lexical items typically occurring in the argument positions. These items, called fillers, form sets belonging to different patterns of meaning.
\namecite{hanks2005} and \namecite{hanks2008} propose an ontology where fillers are clustered into semantic types, i.e. categories such as [[Location]], [[Event]], [[Vehicle]], [[Emotion]]. These form a hierarchy that branches into more specific types.

However, these categories are problematic, as lexical sets  tend to  ``shimmer'' \cite{jezek2010}: their membership tends to change according to the verb they associate with.  The shimmering nature of lexical sets  is not an accidental phenomenon.  Rather, it stems from the fact that verb selectional restrictions may cut across conceptual categories due to the specific predication introduced by the verb. For example, both \textit{wash} and \textit{amputate} typically select [[Body Part]] as their direct object. Nevertheless, they select different prototypical members of the set, as can be seen in the examples below from \namecite{jezek2010} where only shared members are underlined:

\begin{examples}
{[}[Human]] \textit{wash} [[Body Part]]\\
where {[}[Body Part]]: \{\underline{hand}, hair, face, foot, mouth...\}
\end{examples}

\begin{examples}
{[}[Human]] \textit{amputate} [[Body Part]]\\
where {[}[Body Part]]:  \{leg, limb, arm, \underline{hand}, finger...\}
\end{examples}

\noindent Lexical sets can be extracted from corpora automatically. This operation hinges upon traditional techniques for the acquisition of
subcategorisation frames and selectional restrictions. The former allow to capture the syntactic pattern \cite{brent1991automatic,im2000clustering} or semantic frame pattern \cite{baker1998berkeley} in which each verb is found. The latter create probabilistic models of preferences over fillers that are evaluated intrinsically through human judgments \cite{brockmann2003evaluating} and extrinsically through disambiguation tasks \cite{van2014neural}.

\subsection{Causative-Inchoative Verbs}
\label{ssec:cauinc}
In principle, lexical sets can be constructed for every verb. In this work, however, we limit our inquiry to a specific subset of verbs, namely causative-inchoative verbs in Italian. This provides a testbed for our methods of analysis, which can be easily extended to other classes of verbs and alternations. The choice of this specific subset is due to the fact that understanding the internal structure of lexical sets and their relations seems to be crucial to solve the problems surrounding this class of verbs, including asymmetries between transitive objects and intransitive subjects and their relation with the spontaneity scale (see below).

Causative-inchoative verbs alternate. In other terms, they appear in two patterns, either as transitive or intransitive. In the former, an agent brings about a change of state affecting a patient; in the latter, the change of the same patient is presented as spontaneous (e.g.\ {\it break}, as in ``Mary broke the key'' vs. ``the key broke''). The verbs in the two patterns can be expressed by either a same form or two distinct forms cross-lingually. In the second case, the forms can be morphologically asymmetrical: one has a derivative affix and the other does not. Otherwise, the forms are suppletive, being completely unrelated (e.g.\ \textit{kill}/\textit{die}). The members of causative-inchoative verbs that retain a same form or are morphologically related in both patterns vary cross-lingually \cite{montemagni1995}. Alternations regarding physical change-of-state and manner-of-motion are found in English, whereas they are limited to psychological and physical changes-of-state in Italian. In Japanese and Salish languages, also verbs like \textit{arrive} and \textit{appear} do alternate \cite{alexiadou2010}. 

From a semantic point of view, Italian causative-inchoative verbs are required to be telic and have an inanimate patient \cite{cennamo1995}. Morpho-syntactically, they are generated from an asymmetrical derivation, called ``anti-causativisation.'' The intransitive form is sometimes marked with the pronominal clitic \textit{si}: its presence is mandatory, optional or forbidden according to verb-specific rules \cite{cennamo2011}. Because of this, many different categorisations of Italian causative-inchoative verbs were attempted \cite{folli2002,jezek2003}.

Causative-inchoative verbs in general are endowed with peculiar properties. \namecite{haspelmath1993} claims that verbs with a cross-lingual preference for a marked causative form denote a more ``spontaneous'' situation. Spontaneity is intended by the author as the likelihood that the occurrence of the event denoted by the verb does not require the intervention of an agent. In this way, a correlation between the form and the meaning of these verbs was borne out. Moreover, \namecite{samardzic2012meaning} and \namecite{haspelmath2014coding} demonstrated that verbs appearing more frequently (intra- and cross-lingually) in the inchoative form tend to morphologically derive the causative form. Here, the correlation bridges between form and frequency. Vice versa, situations entailing an agentive participation prefer to mark the inchoative form and occur more frequently in the causative form.

However, what remains uncertain is whether spontaneous and agentive variants of the same verbs differ in their lexical sets. \namecite{atkins1995building} and \namecite{levin1995unaccusativity} argued that selectional restrictions for spontaneous verbs (named internal causation verbs) are stricter because their event unfolds due to some inherent property of the patient: referents without this capability are excluded, contrary to what happens with agentive verbs (named external causation verbs). This capability is defined ``teleological'' by \cite{copley2014theories}. However, \namecite{mckoon2000} reported contradicting results from corpus-based analyses that did not find any significant difference in the content of lexical sets of spontaneous and agentive verbs (although from a sample of less than 100 sentences and only 5 categories). Instead, they reported a difference between transitive objects and intransitive objects in that the former contained a larger amount of abstract nouns compared to the latter.


\subsection{Distributional Semantics}
\label{ssec:we}
Once established the domain, we need to provide a reliable method of inquiry. Previous works based on set theory treated lexical sets as Aristotelian categories, of which a filler is either a member or not. For instance, \namecite{montemagni1995} collected lexical sets manually and employed set intersection as a measure of similarity between two sets. Research in psychology, however, has long since demonstrated that the members of a linguistic set are found in a radial continuum where the most central element is the prototype for its category, and those at the periphery are less representative \cite{rosch1973,lakoff1987}.\footnote{For previous work on lexical sets considering prototypicality in the context of the notion of shimmering, see \namecite{jezek2010}.}

The full exploitation of the semantic information inherent to the argument fillers of verbs can take advantage of some recent developments in distributional semantics. Efficient algorithms have been devised to map each word of a vocabulary into a corresponding vector of \textit{n} real numbers, which can be thought as a sequence of coordinates in a \textit{n}-dimensional space \cite[\textit{inter alia}]{mikolov2013}. This mapping is yielded by unsupervised machine learning, based on the assumption that the meaning of a word can be inferred by its context, i.e.\ its neighbouring words in texts. This is known as Distributional Hypothesis \cite{harris1954distributional,firth1957synopsis}. Distributed models have some relevant properties: first, the geometric closeness of two vectors corresponds to the similarity in meaning of the corresponding words. Moreover, its dimensions possibly retain a semantic interpretation such that non-trivial analogies can be established among words.

Word vectors allow to capture the spatial continuum implied by the notion of prototype. Previous works showed that word vectors can be clustered to imitate linguistic categories, because each cluster captures the `semantic landscape' of a word \cite{hilpert2015meaning}. The center of these clusters can be interpreted as the prototype of the corresponding category, and the proximity of the cluster members to the center as the degree of their prototypicality. In fact, the cluster members are not scattered randomly, but rather are arranged according to the internal structure of the cluster \cite{dubossarsky2016verbs}. The prototypicality of the cluster members provides an explanation about linguistic phenomena, such as the resistance to the diachronic change of meaning \cite{geeraerts1999diachronic,dubossarsky2015bottom}.

In this work, we extend the usage of the notion of prototypicality from meaning-based categories derived through vector quantization to grammatically defined categories (i.e.\ lexical sets) derived through dependency parsing. Moreover, we go beyond the estimation of the distance of each word from the centroid of its category: in particular, we propose new methods to assess the internal diversity in terms of sub-categories and the distance between lexical sets themselves.

\begin{table}[b!]
\rowcolors{1}{white}{LighterGray}
\centering
\caption{List of 20 causative-inchoative verbs and count of their fillers for each argument slot.}
\label{tab:verblist}
\begin{tabularx}{0.8\linewidth}{XX cc}
\toprule
\textbf{Lemma} & \textbf{Translation} & S & O \\ 
\hline
chiuder(si) & to close & 289 & 606\\
aprir(si) & to open & 195 & 1337\\
aumentare & to improve & 534 & 998\\
romper(si) & to break & 83 & 419\\
riempir(si) & to fill &58 & 166\\
raccoglier(si) & to gather& 85 & 505\\
connetter(si) & to connect& 39 & 134\\
divider(si) & to split  & 129 & 246\\
finire & to stop & 1092 & 721\\
uscire/portare fuori & to go/put out & 325 & 638\\
alzar(si) & to arise/raise & 75 & 304\\
scuoter(si) & to rock & 10 & 69\\
bruciare & to burn & 75 & 174\\
congelare & to freeze & 10 & 30\\
girare & to spin & 155 & 243\\
seccare & to dry & 15 & 14\\
svegliar(si) & to awake/wake & 68 & 89\\
scioglier(si) & to melt & 94 & 143\\
(far) bollire & to boil & 2 & 2\\
affondare & to sink & 18 & 74\\
\hline
\end{tabularx}
\end{table}

\section{Data and Method}
\label{sec:data}

We sourced the lexical sets from a sample of ItWac, a wide corpus gathered by crawling texts from the Italian domain in the web using medium frequency vocabulary as seeds \cite{baroni2009}. This sample was further enriched with morpho-syntactic information through the graph-based \textsc{mate}-tools parser \cite{bohnet2010}. We trained and evaluated this parser on the Italian treebank inside the collection of Universal Dependencies v1.3 \cite{nivre2016universal}. The evaluation on gold standard data suggests how many errors we expect to affect the predictions on the new data, i.e.\ the ItWac corpus: these errors are then propagated to the following steps for the extraction of lexical sets. According to the LAS metric, the relevant dependency relations scored 0.751 for \textit{dobj} (direct object), 0.719 for \textit{nsubj} (subject), and 0.691 for \textit{nsubjpass} (subject of a passive verb). A target group of 20 causative-inchoative verbs was taken from \namecite{haspelmath2014coding}: they are listed in Table \ref{tab:verblist}, together with the count of the extracted lexical sets for the relevant semantic macro-roles (see below).

Once the sentences were parsed, the target verbs were identified inside the dependency trees. The lemmas of these verbs and the forms of their arguments were stored in a database. Argument fillers were grouped according to the semantic macro-roles defined by \namecite{dixon1994}, rather than their dependency relations: subjects of transitive verbs (A), subjects of intransitive verbs (S) and objects (O). In particular, the subjects of verb forms accompanied by the \textit{si}-clitic and those without an object depending on the same verb were treated as S.\footnote{Subjects of verbs inflected in the passive voice were treated as O, instead.} These operations resulted in a database structured as follows: in each row, a verb is alongside of the fillers it occurs with in a specific sentence. For example, compare an original sentence and its corresponding entry in Example \ref{ex3}:

\begin{examples}
\gll \textbf{Plinio} il Vecchio non \textbf{cita} pi\`u il \textbf{Po} di Adria perch\'e l' \textbf{argine} dell' Adige si era \textbf{rotto} ed era \textbf{confluito} nella Filistina.
Pliny the Elder doesn't mention anymore the Po of Adria because the bank {of the} Adige {} had broken and had merged {with the} Filistina.
\glt
\glend
\label{ex3}
\end{examples}
\begin{table}[h!]
\rowcolors{1}{white}{LighterGray}
\centering
\begin{tabularx}{0.5\linewidth}{X XXX}
\toprule  \textbf{Verb} &	\textbf{A} &	\textbf{S} &	\textbf{O} \\ 
\hline  citare &	Plinio		& \_ & Po \\ 
rompere &  & argine & \\
confluire & \_ & \_ & \_ \\
\hline 
\end{tabularx}
\end{table}

Because of the nature of vector models, we made the following design choices to deal with special linguistic phenomena. We discarded everything but the head of multi-word nouns, such as \textit{Plinio} of \textit{Plinio il Vecchio}, to preserve the one-to-one mapping between words and vectors. We did not distinguish proper and common nouns, such as \textit{Po} and \textit{argine}, since their representations lie in the same multi-dimensional space. Subjects in ellipsis or co-reference were neglected, since no pragmatic annotation of the sentences was available: for instance, \textit{Adige} should appear as S in the entry for \textit{confluire}, which is left blank instead. Finally, polysemous words and homonyms were represented by a single form, hence a single vector, since their representations conflate all the relevant meanings. For instance, \textit{Adige} is both a river and a location, but these are not distinguished in the vector model. 

The database was later collapsed by verb lemma so that each of them became associated to three sets of fillers (one per macro-role). Each of these sets is a corpus-based lexical set. Compared to manually picked lexical sets, they are more noisy but less sparse: the vastness of the data mitigates the errors in the parsing step. Moreover, the automation in lexical set extraction allows to access the fillers of virtually every verb: resources based on manual selection like T-PAS \cite{jezek2014}, on the other hand, are limited to a small amount of verbs.

Afterwards, each of the argument fillers was mapped to a vector according to three different pre-trained models. The vectors are generated unsupervisedly  back-propagating the gradient from the loss of a task to update randomly initialized embeddings. Each model, however, relies on different tasks:

\begin{itemize}
\justifying
\item \textbf{CBOW} stands for Continuous Bag of Words and is part of the Word2Vec suite \cite{mikolov2013b}. The Italian model was developed by \namecite{dinu2015} through negative sampling. 300-dimensional representations were obtained by training a binary classifier that discriminates whether a pair of a target word \textit{w} and a context \textit{c} belongs to an actual text or not. True contexts are obtained from a window of 5 words on either side of the target word, false contexts are drawn randomly from a noise distribution. Moreover, infrequent words are pruned. On the other hand, a preprocessing step called subsampling deletes words from windows whose frequency \textit{f} is higher than an threshold with a probability $p = 1- \sqrt{\frac{t}{f}}$. The text from which word-context pairs are sampled is the full ItWac corpus, consisting in 1.6 billion tokens. The representations resulting from this algorithm are organised by topical similarity or relatedness.
\item \textbf{fastText} embeddings were trained on the dump of the Italian Wikipedia through a character-level skip-gram \cite{bojanowski2016enriching}. The skip-gram algorithm predicts the following element given a sequence of previous elements. The elements of Bojanowski's version are characters: each word is represented by a bag of character n-grams and its representation consists in the sum of the representations of each of them. The vector dimensionality is 300; random sequences are drawn with a ratio of 5 to 1 compared to true sequences, with a probability proportional to the square root of the uni-gram frequency.
The size of the context window is uniformly sampled from values between 1 and 5. The rejection threshold for subsampling is $10^{-4}$. The usage of character n-grams makes this vector model sensitive to morphological similarity.
\item \textbf{Polyglot} vectors result from a classifier distinguishing between sequences taken from the dump of the Italian Wikipedia and corrupted sequences \cite{polyglot:2013:ACL-CoNLL}. The window of phrases was 5, the dimensionality of vectors 64. Subsampling discarded words not appearing in the raking of the 100 thousand most frequent ones.
\end{itemize}

In order to measure spatial distances inside these vector models, many different metrics are available, including pure geometrical (Euclidean) distance. In this work, we rely on the popular metrics of cosine distance \cite{cha2007comprehensive}. Assume that \textbf{a} and \textbf{b} are vectors, and that \textbf{a}$_i$ and \textbf{b}$_i$ are their $i^{th}$ components, respectively. The cosine similarity between \textbf{a} and \textbf{b} is then defined as follows:

\begin{equation}
\cos(\theta) = \frac{\sum_{i=1}^{n}\textbf{a}_i\textbf{b}_i}{\sqrt{\sum_{i=1}^{n}\textbf{a}_i^2}\sqrt{\sum_{i=1}^{n}\textbf{b}_i^2}}
\end{equation}

The opposite, namely cosine distance \textit{d}, is simply defined as $1 - \cos(\theta)$. As for the values that \textit{d}(\textbf{a},\textbf{b}) can assume, the minimum is at 0 (angles completely overlap) and the maximum is at 1 (orthogonal vectors). 

\section{Experiments}
\label{sec:experiments}

\begin{figure*}[ph!]
     \begin{center}
        \subfigure[chiudere]{%
            \label{fig:first}
            \includegraphics[width=0.24\textwidth]{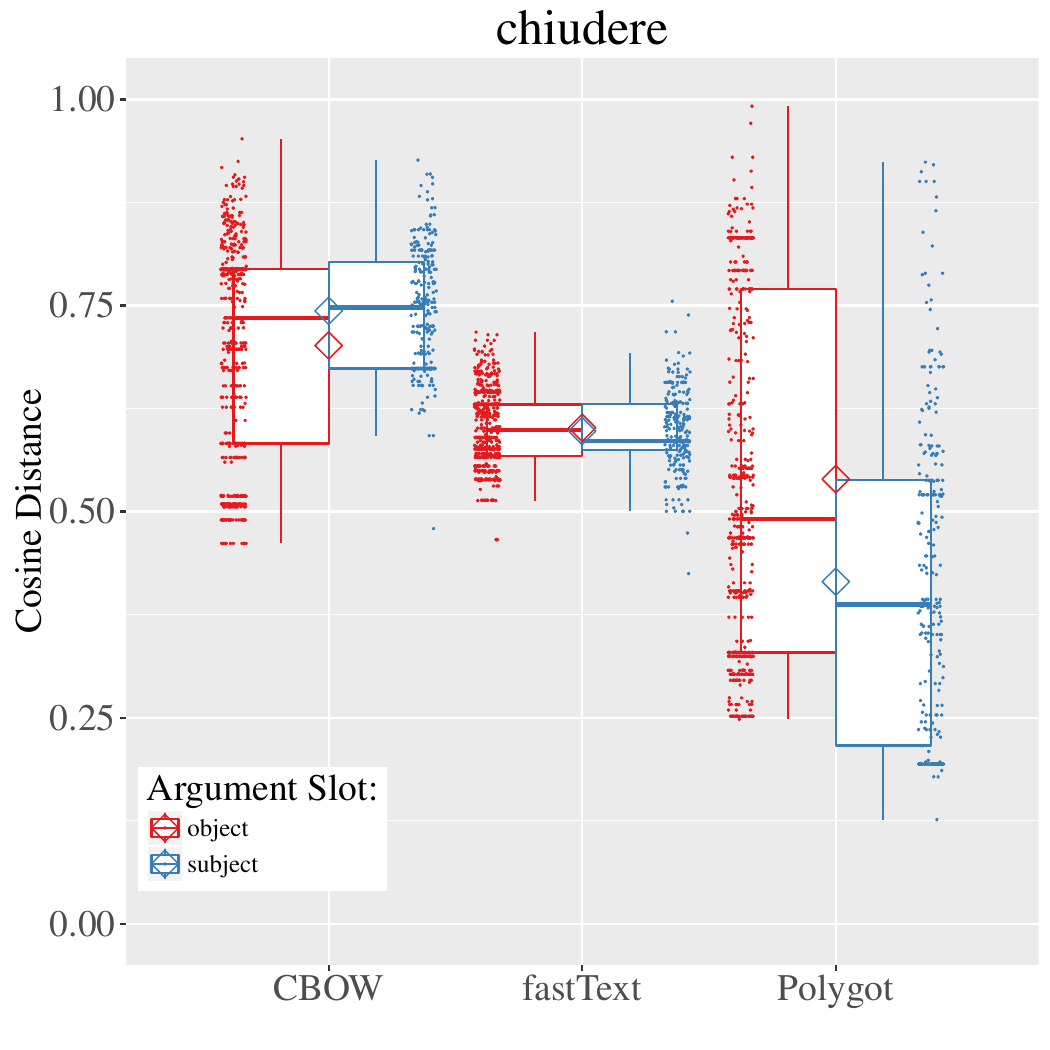}
        }%
        \subfigure[aprire]{%
           \label{fig:second}
           \includegraphics[width=0.24\textwidth]{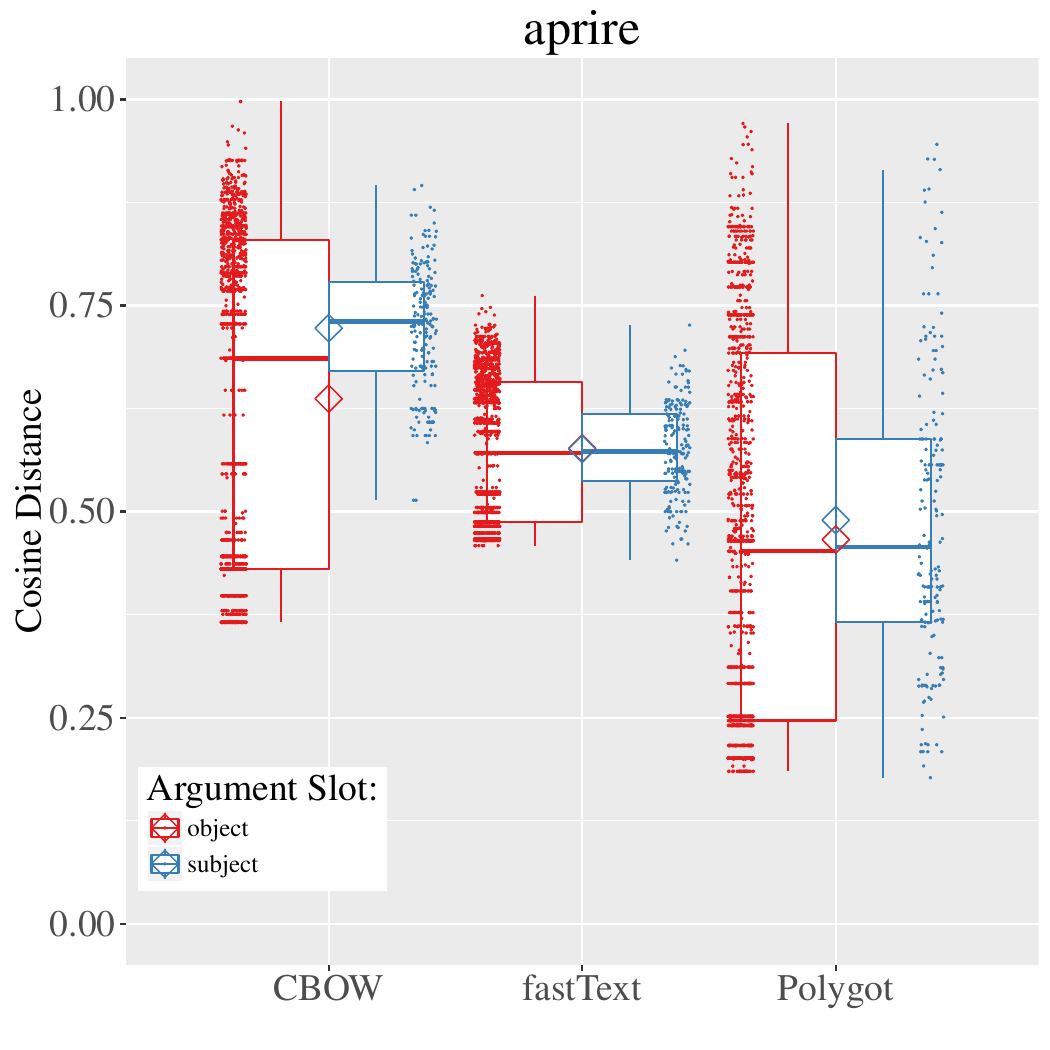}
        }%
        \subfigure[aumentare]{%
            \label{fig:third}
            \includegraphics[width=0.24\textwidth]{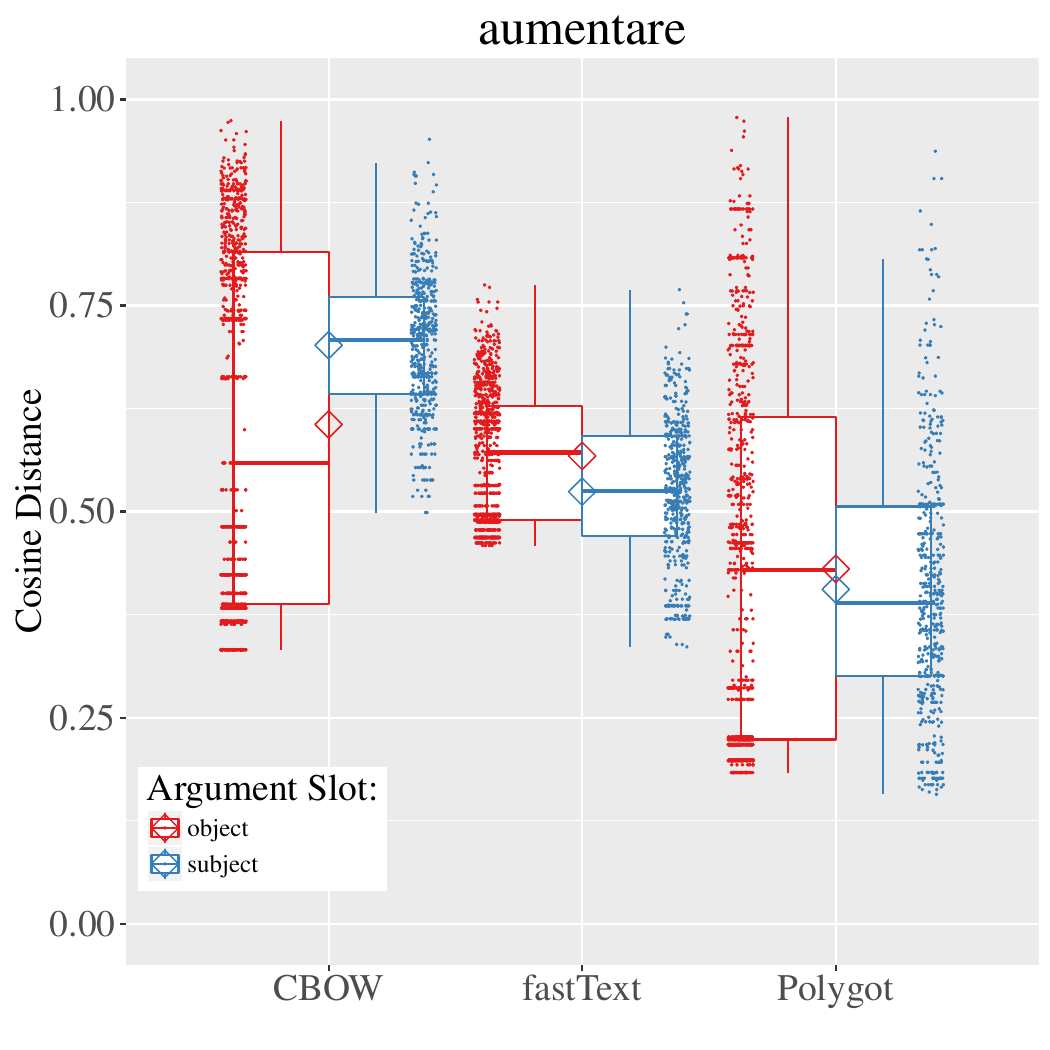}
        }%
        \subfigure[rompere]{%
           \label{fig:fourth}
           \includegraphics[width=0.24\textwidth]{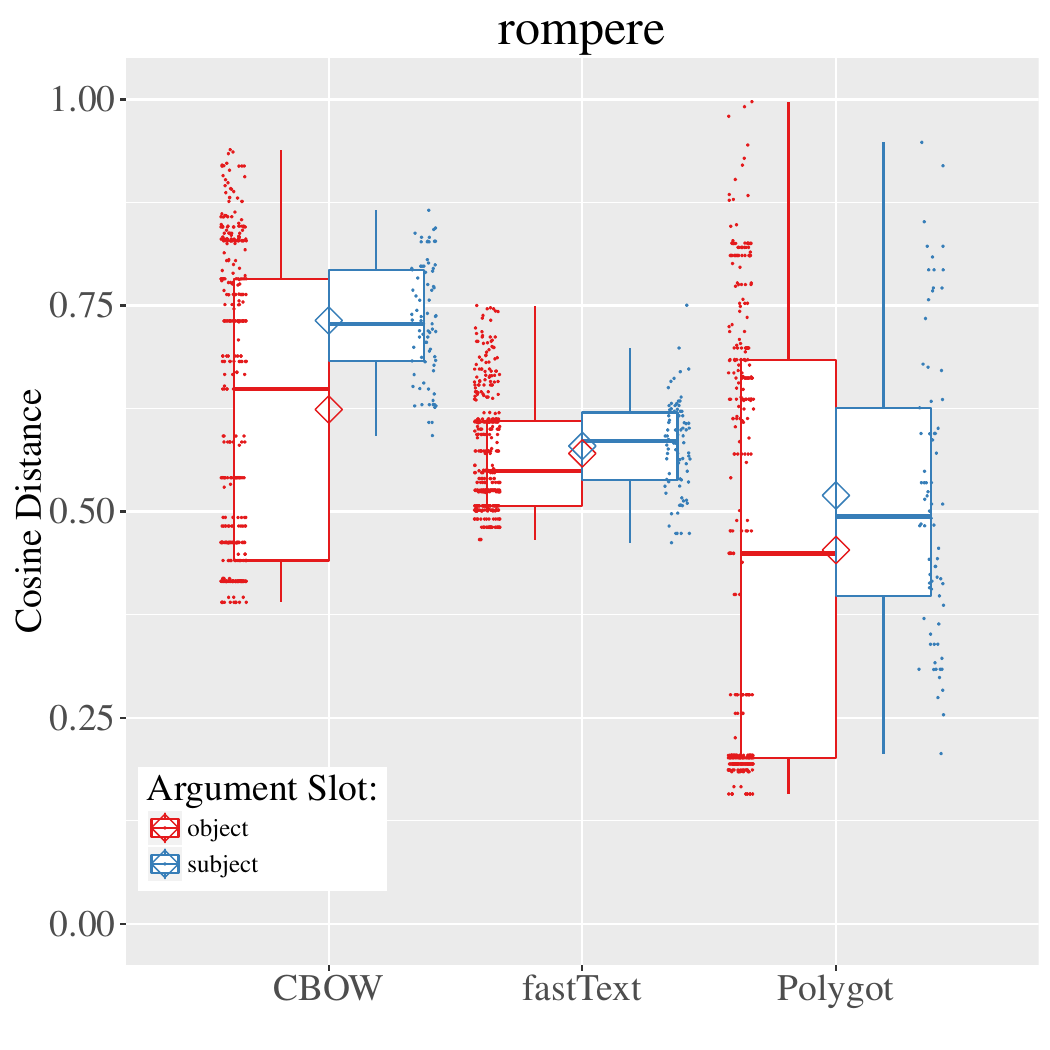}
        }%
\\ 
\vspace{-0.5\baselineskip}
        \subfigure[riempire]{%
            \label{fig:fifth}
            \includegraphics[width=0.24\textwidth]{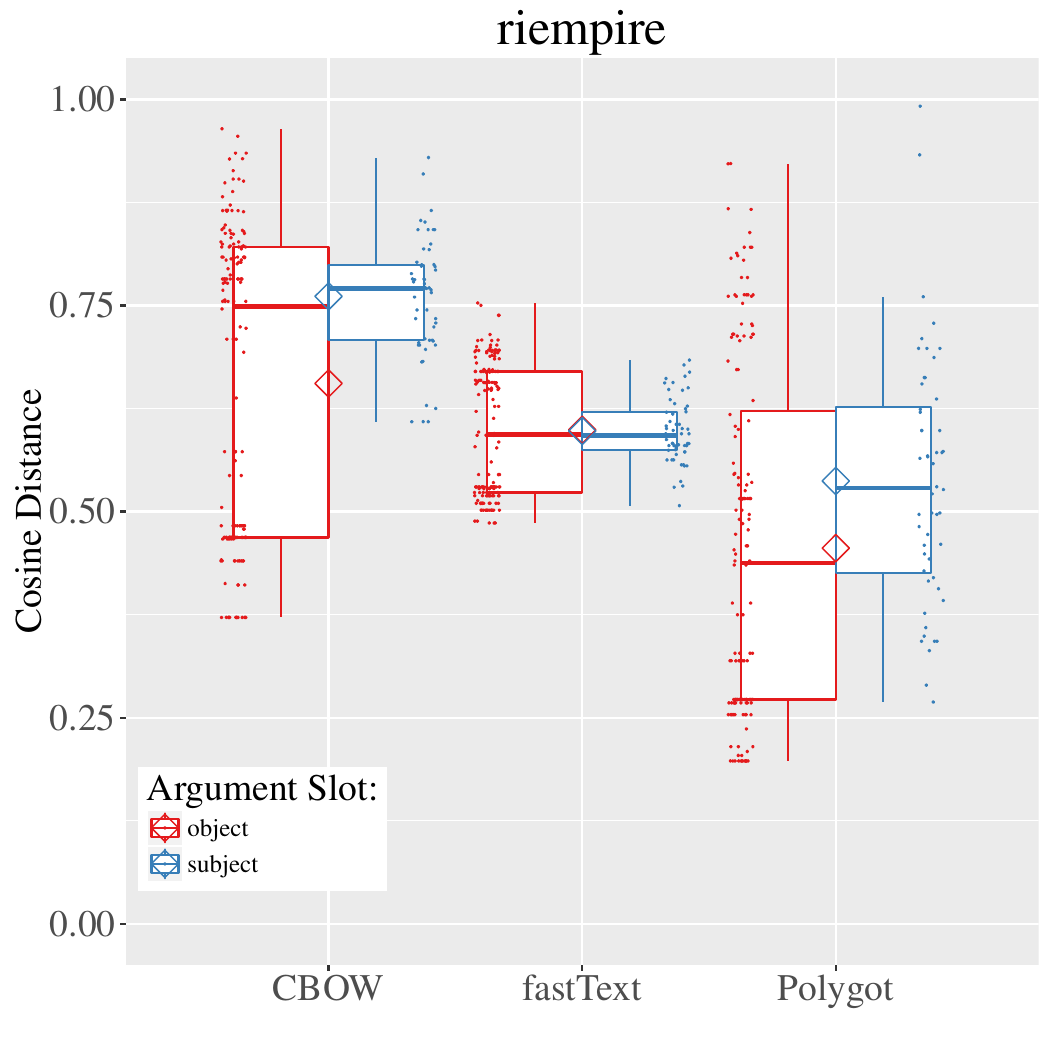}
        }%
        \subfigure[raccogliere]{%
           \label{fig:sixth}
           \includegraphics[width=0.24\textwidth]{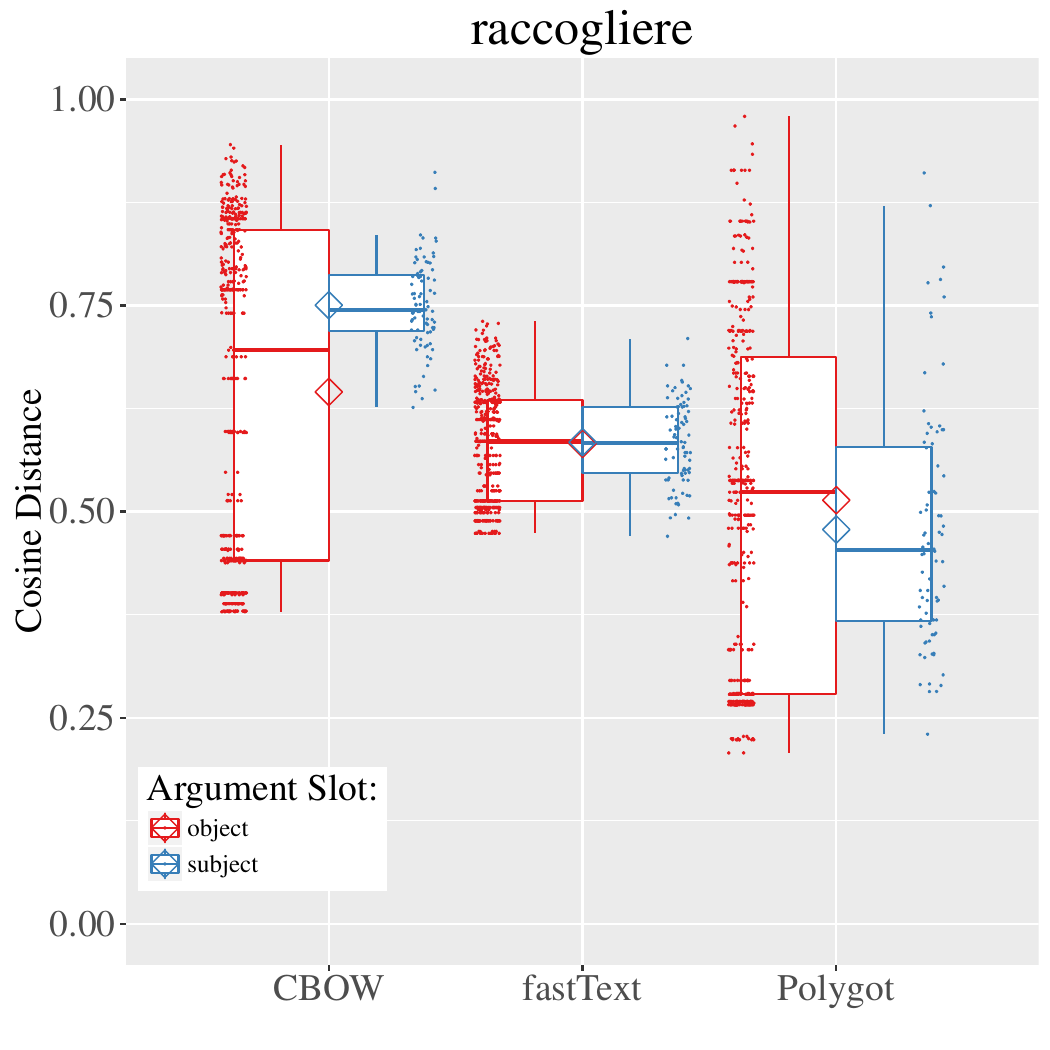}
        }%
        \subfigure[connettere]{%
            \label{fig:seventh}
            \includegraphics[width=0.24\textwidth]{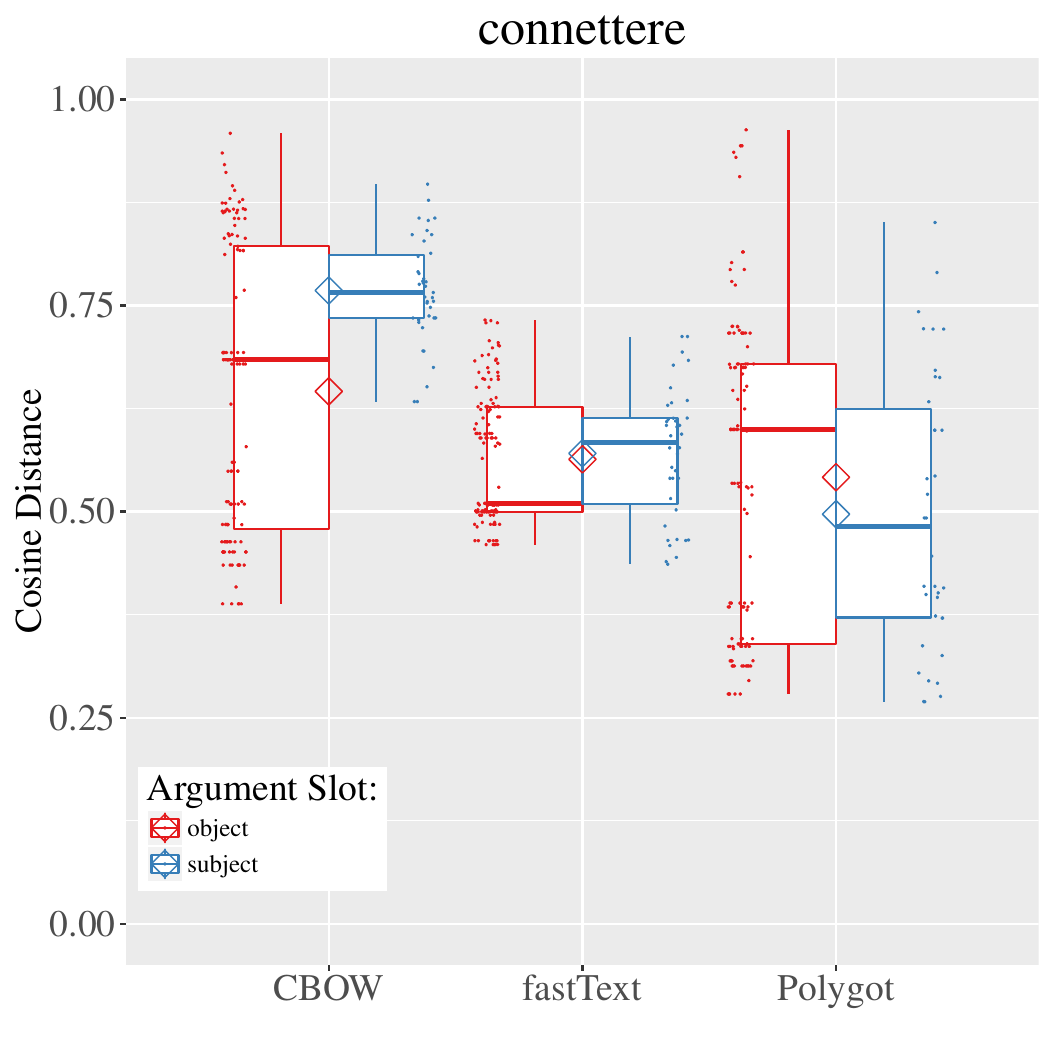}
        }%
        \subfigure[dividere]{%
           \label{fig:eigth}
           \includegraphics[width=0.24\textwidth]{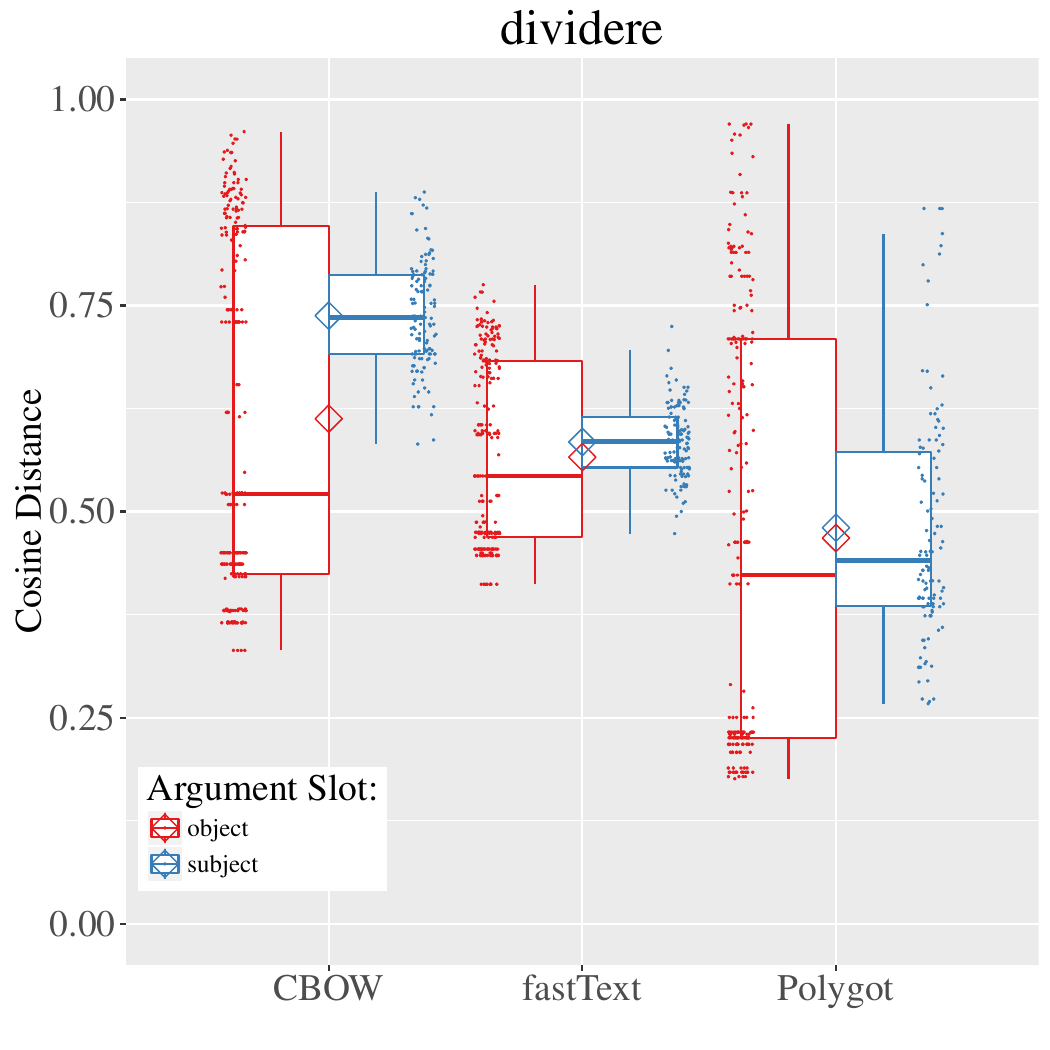}
        }%
\\ 
\vspace{-0.5\baselineskip}
        \subfigure[finire]{%
            \label{fig:ninth}
            \includegraphics[width=0.24\textwidth]{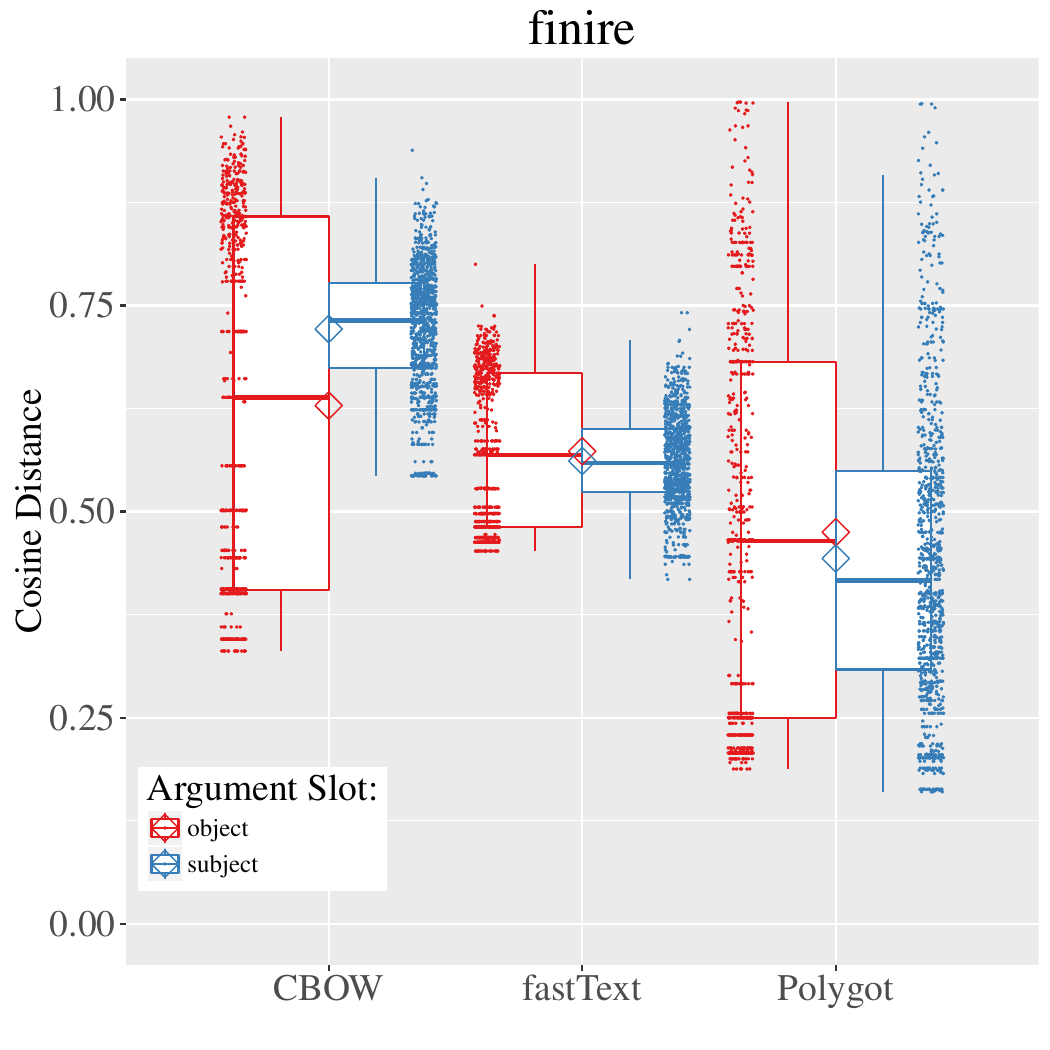}
        }%
        \subfigure[uscire]{%
           \label{fig:tenth}
           \includegraphics[width=0.24\textwidth]{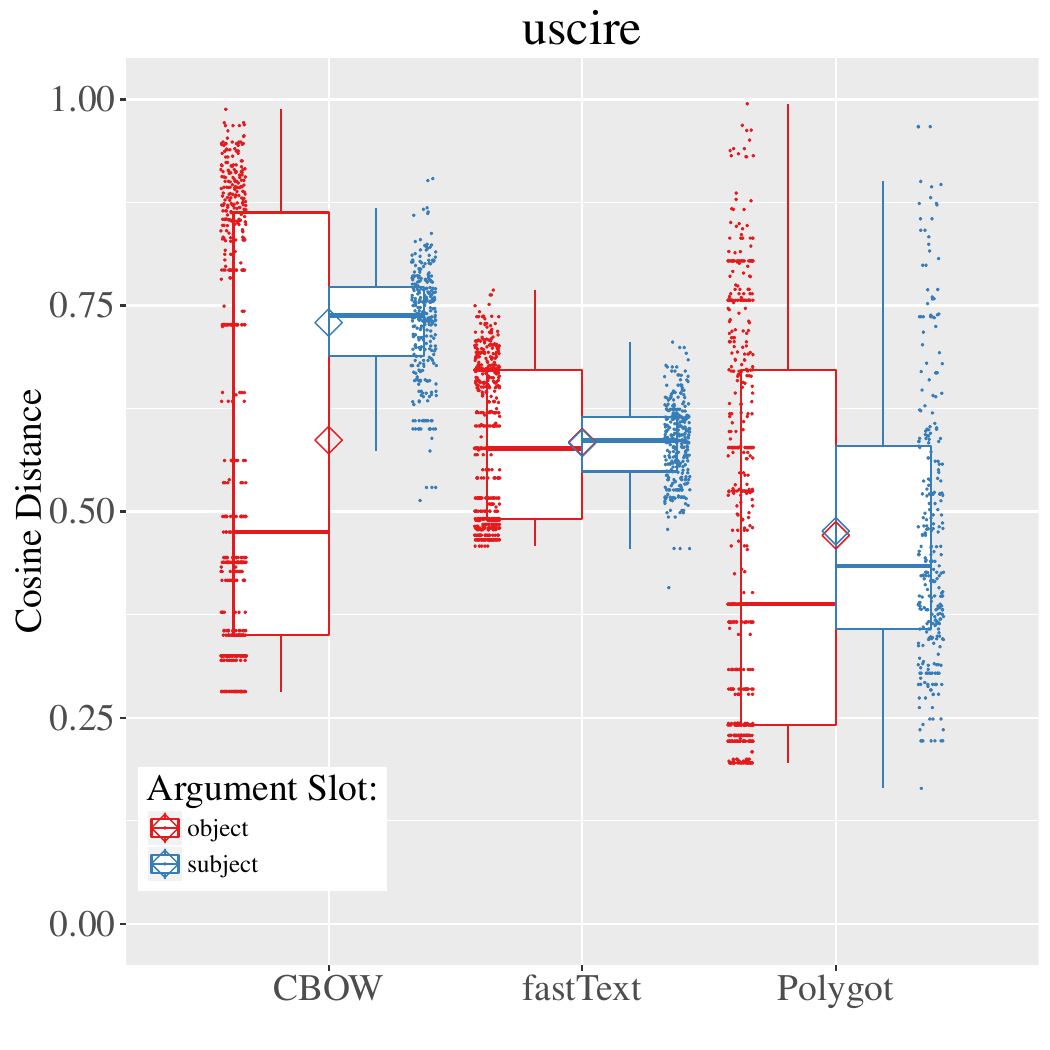}
        }%
        \subfigure[alzare]{%
            \label{fig:11th}
            \includegraphics[width=0.24\textwidth]{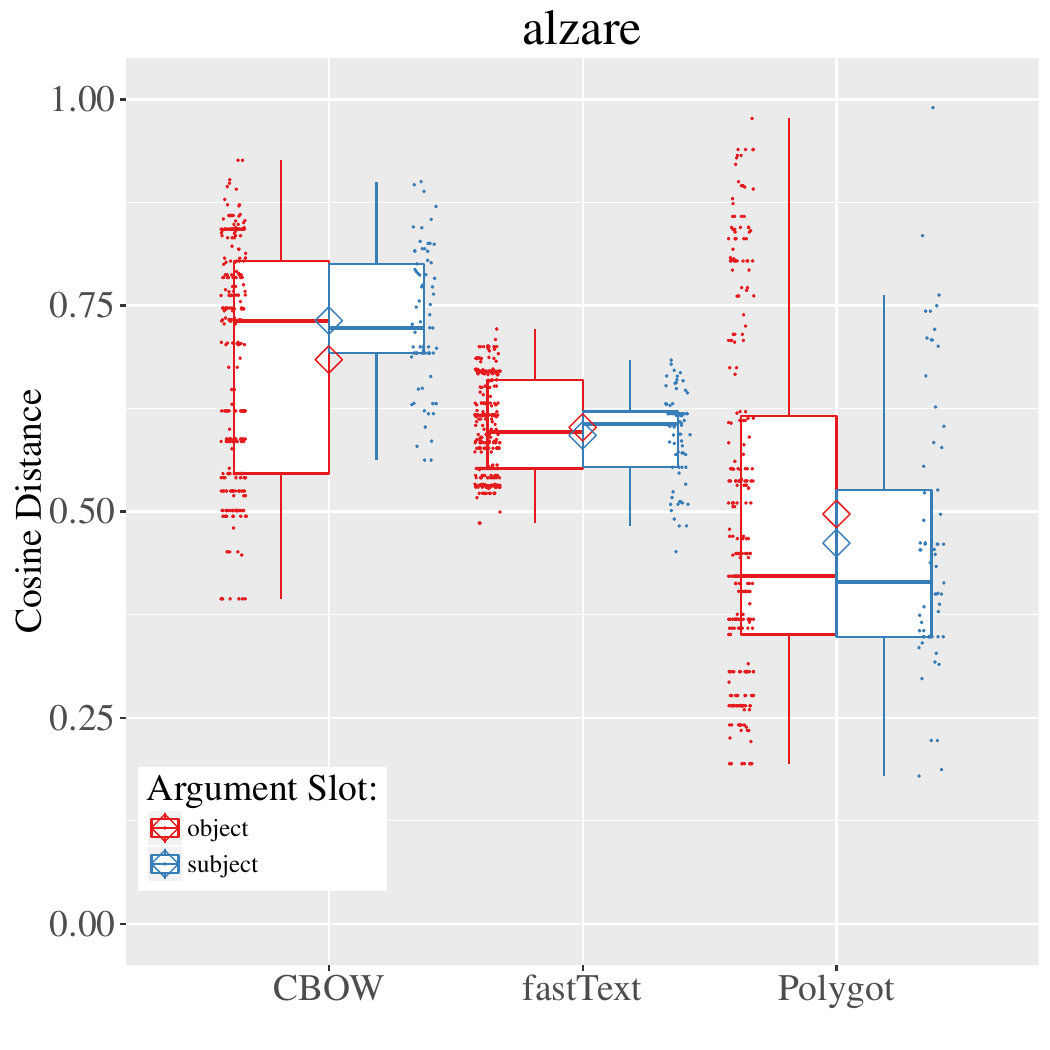}
        }%
        \subfigure[scuotere]{%
           \label{fig:12th}
           \includegraphics[width=0.24\textwidth]{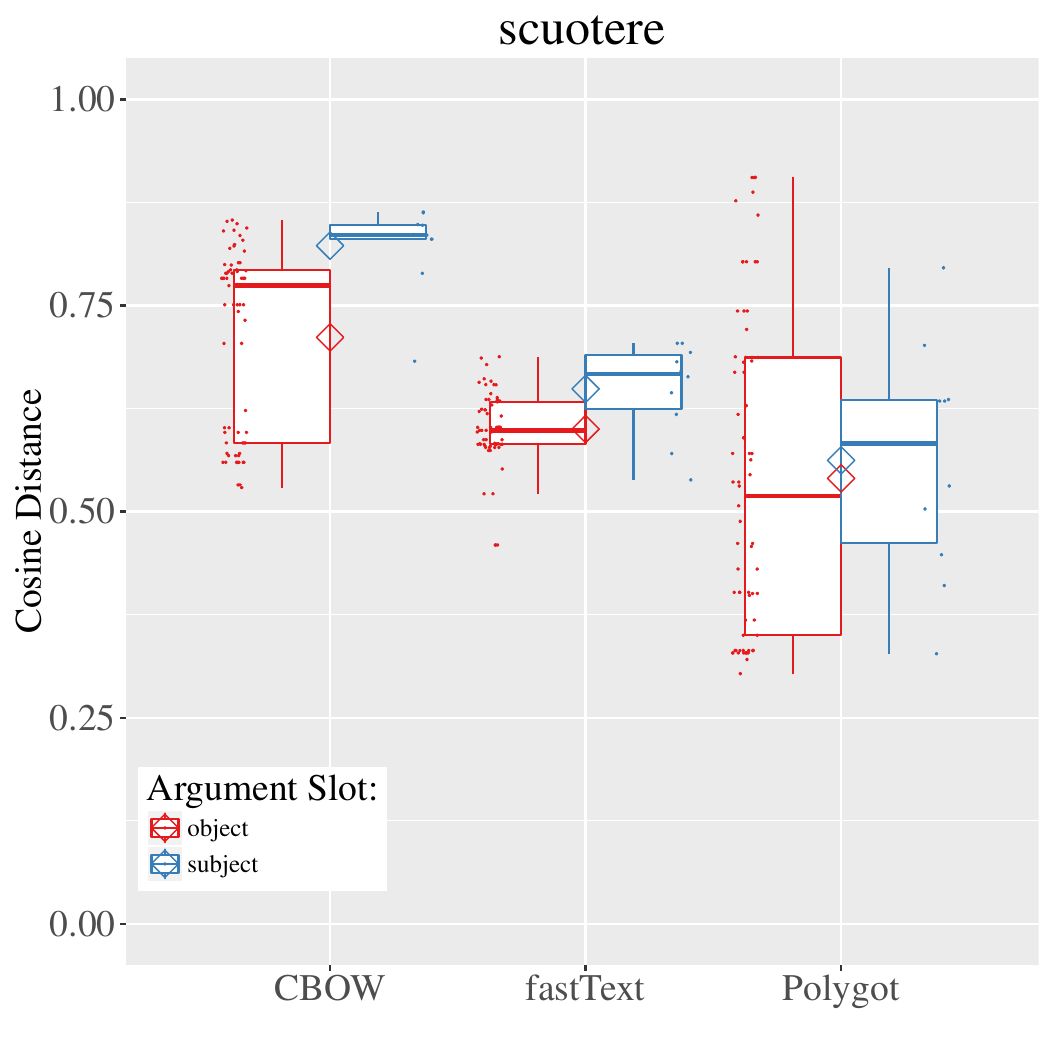}
        }%
\\ 
\vspace{-0.5\baselineskip}
        \subfigure[bruciare]{%
            \label{fig:13th}
            \includegraphics[width=0.24\textwidth]{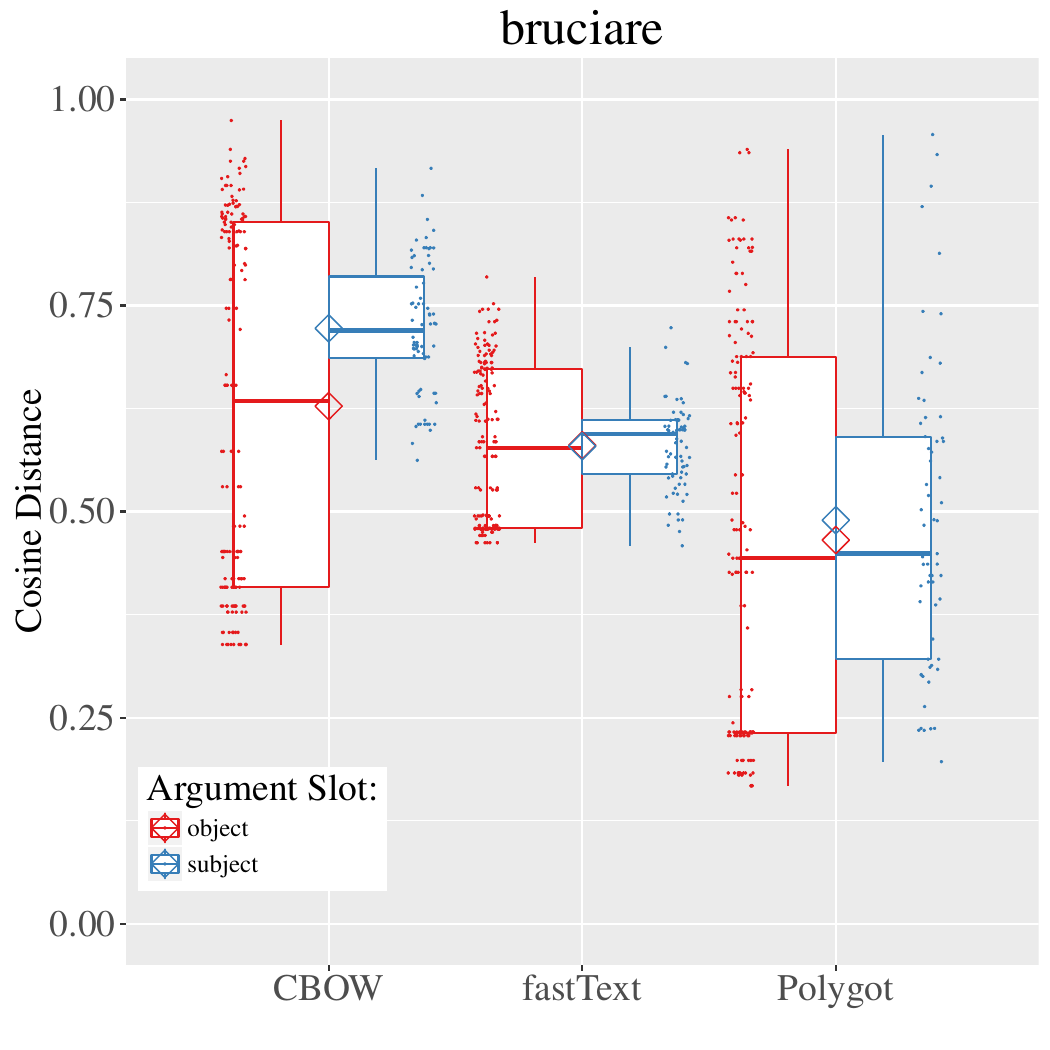}
        }%
        \subfigure[congelare]{%
           \label{fig:14th}
           \includegraphics[width=0.24\textwidth]{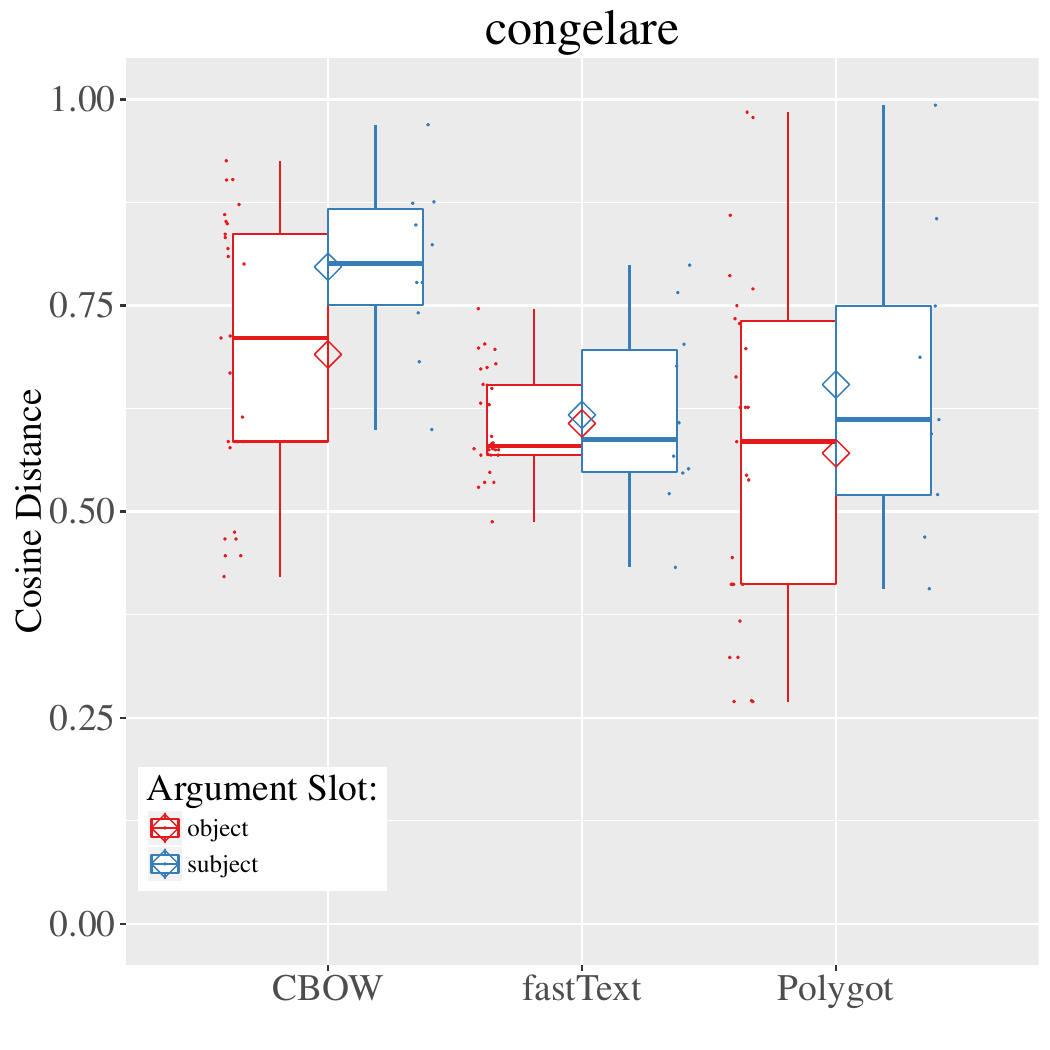}
        }%
        \subfigure[girare]{%
            \label{fig:15th}
            \includegraphics[width=0.24\textwidth]{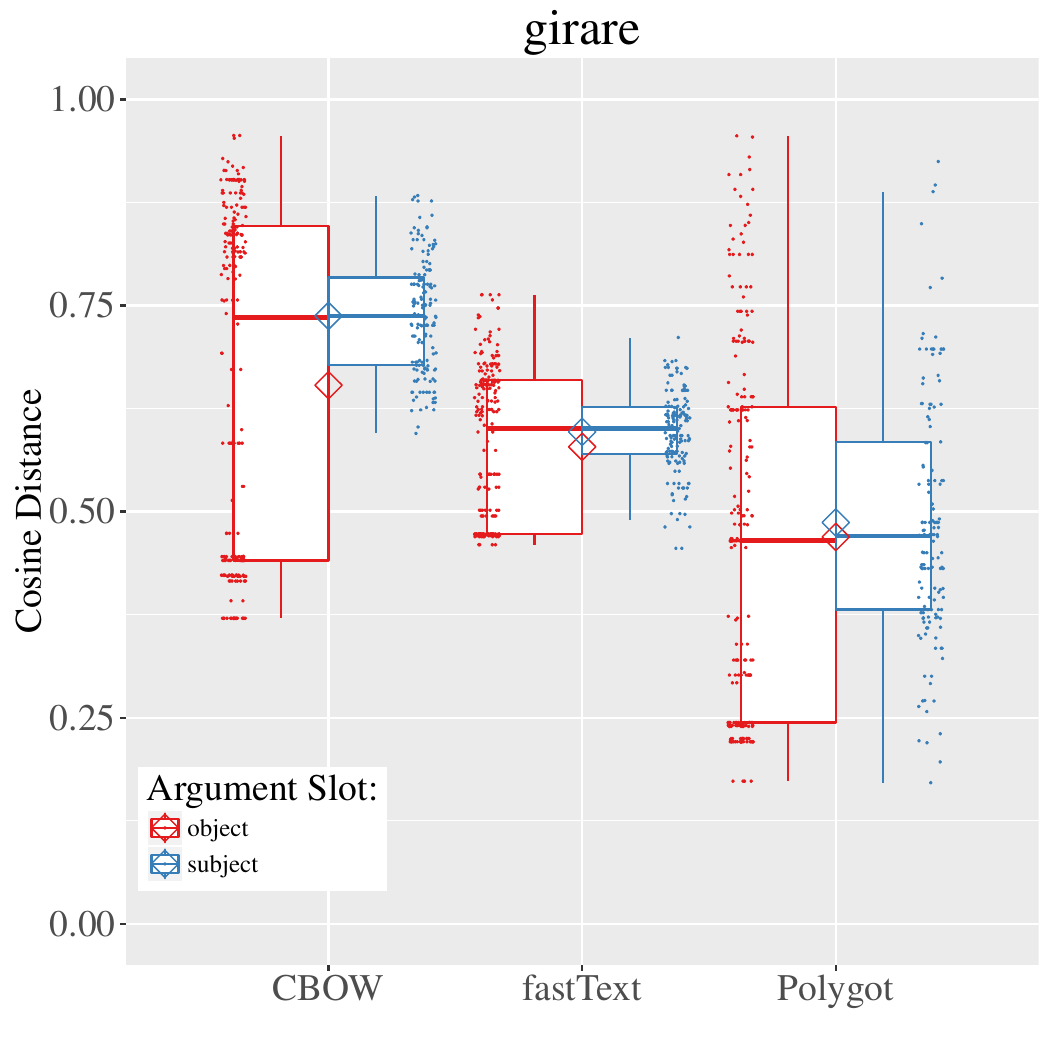}
        }%
        \subfigure[seccare]{%
           \label{fig:16th}
           \includegraphics[width=0.24\textwidth]{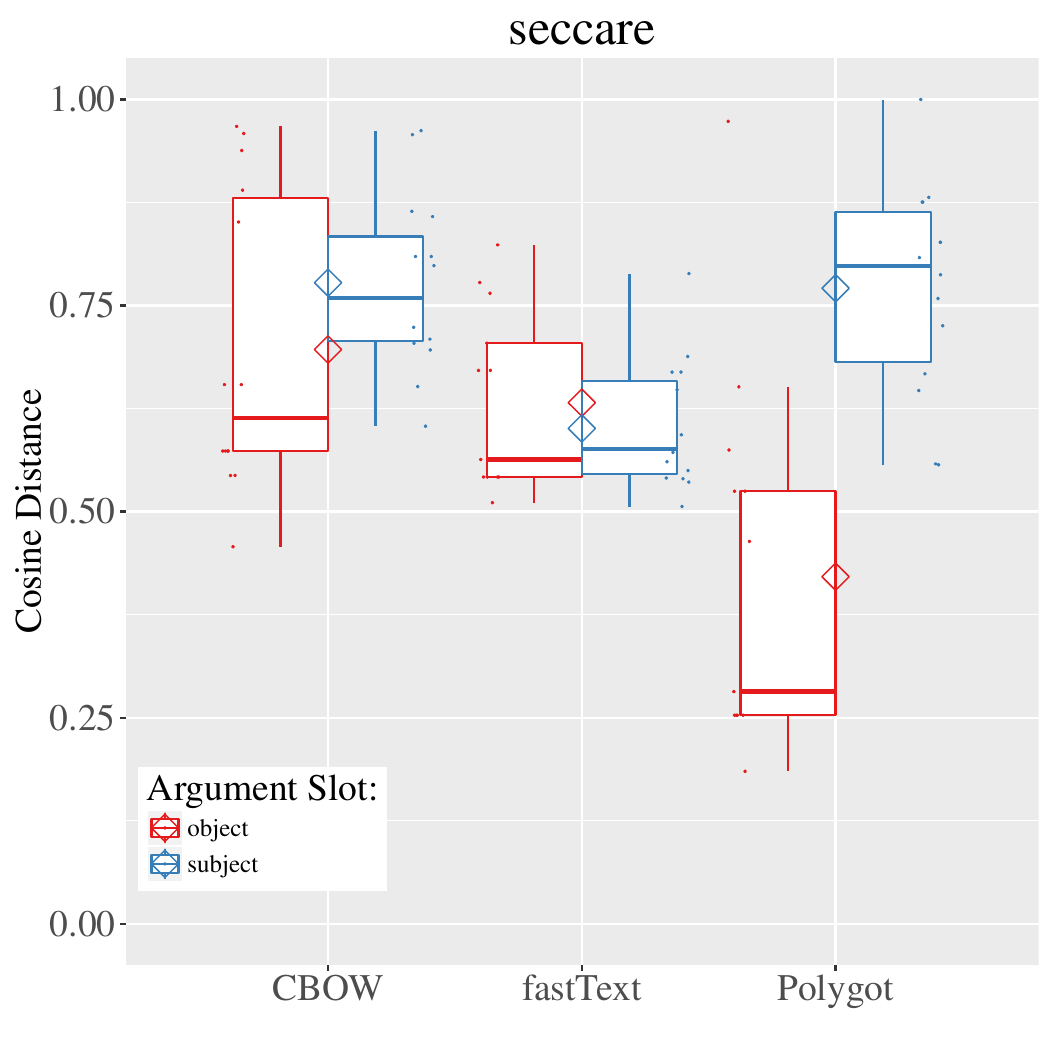}
        }%
\\ 
\vspace{-0.5\baselineskip}
        \subfigure[svegliare]{%
            \label{fig:17th}
            \includegraphics[width=0.24\textwidth]{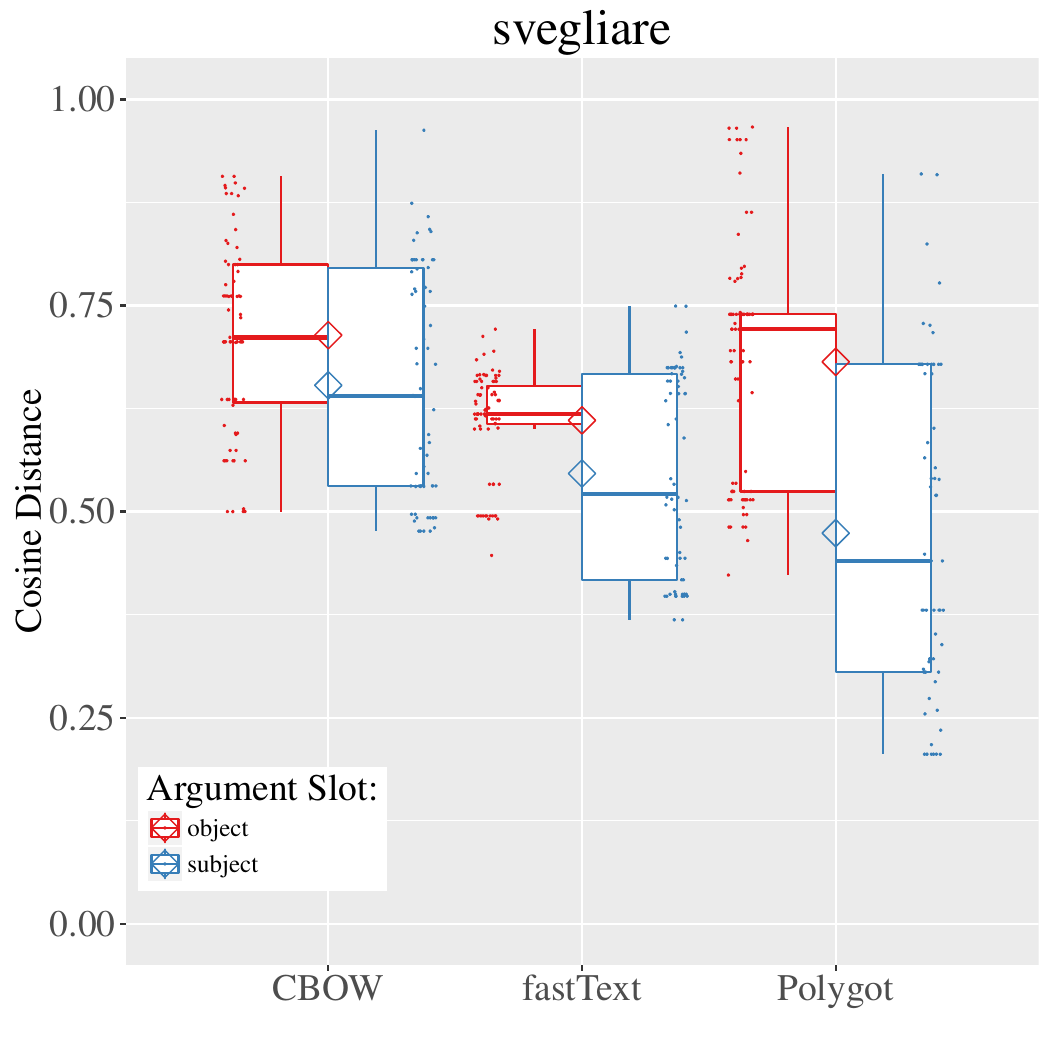}
        }%
        \subfigure[sciogliere]{%
           \label{fig:18th}
           \includegraphics[width=0.24\textwidth]{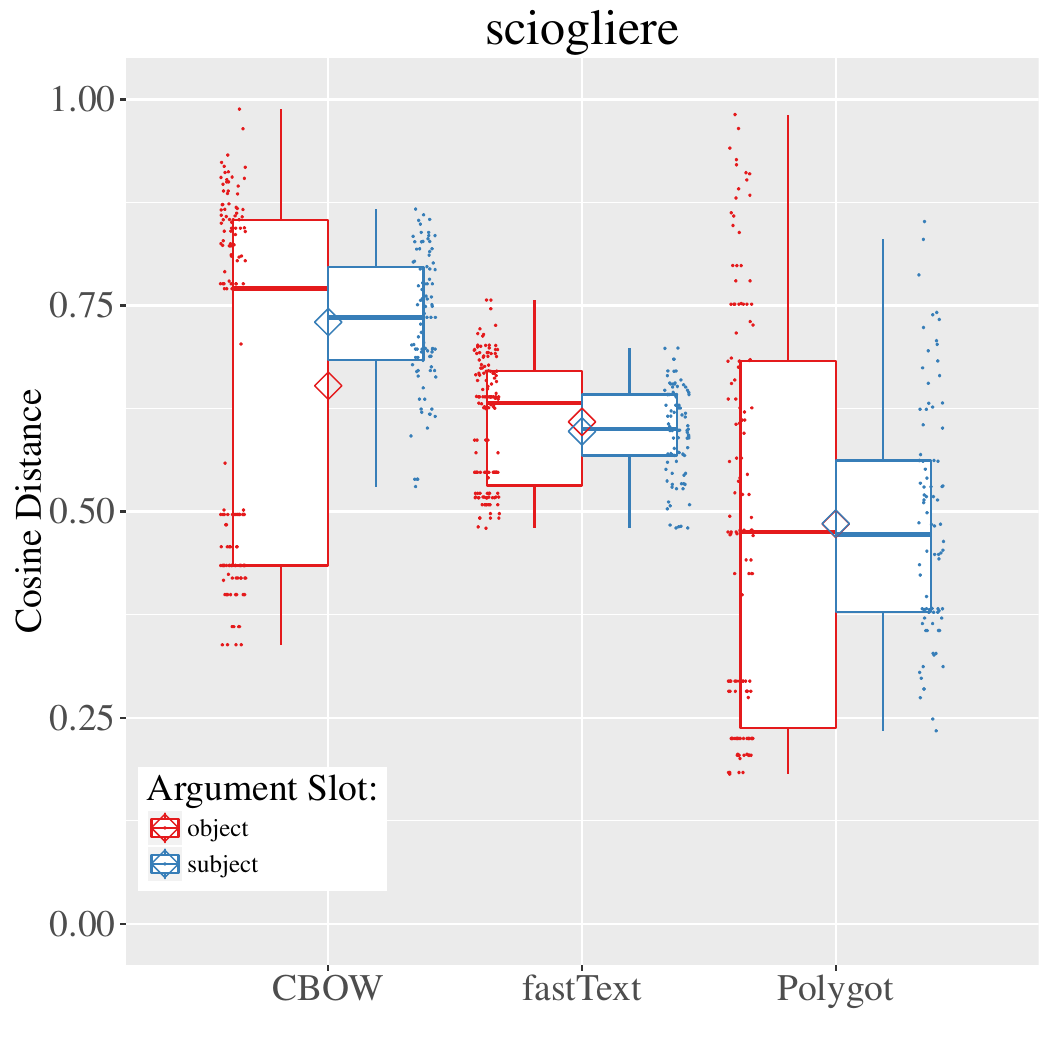}
        }%
        \subfigure[bollire]{%
            \label{fig:19th}
            \includegraphics[width=0.24\textwidth]{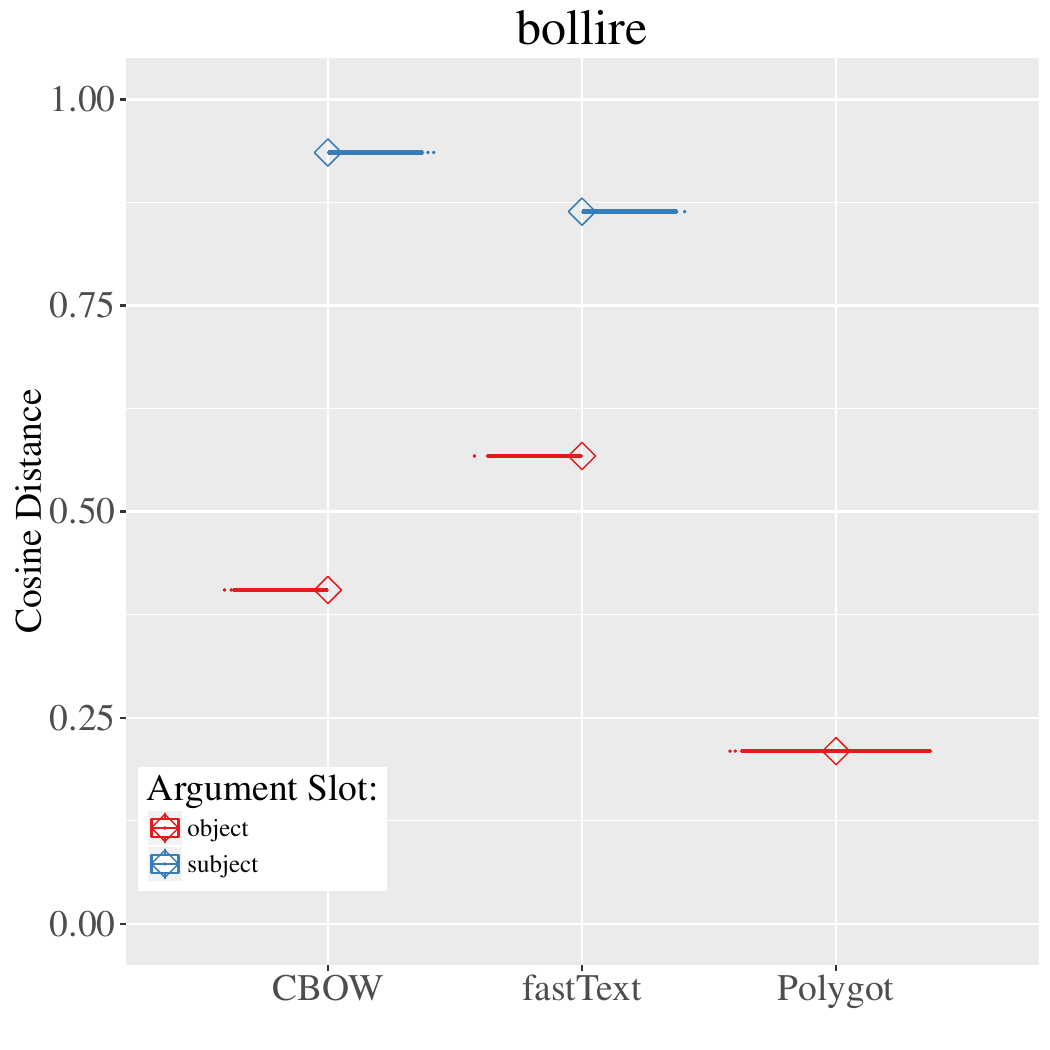}
        }%
        \subfigure[affondare]{%
           \label{fig:20th}
           \includegraphics[width=0.24\textwidth]{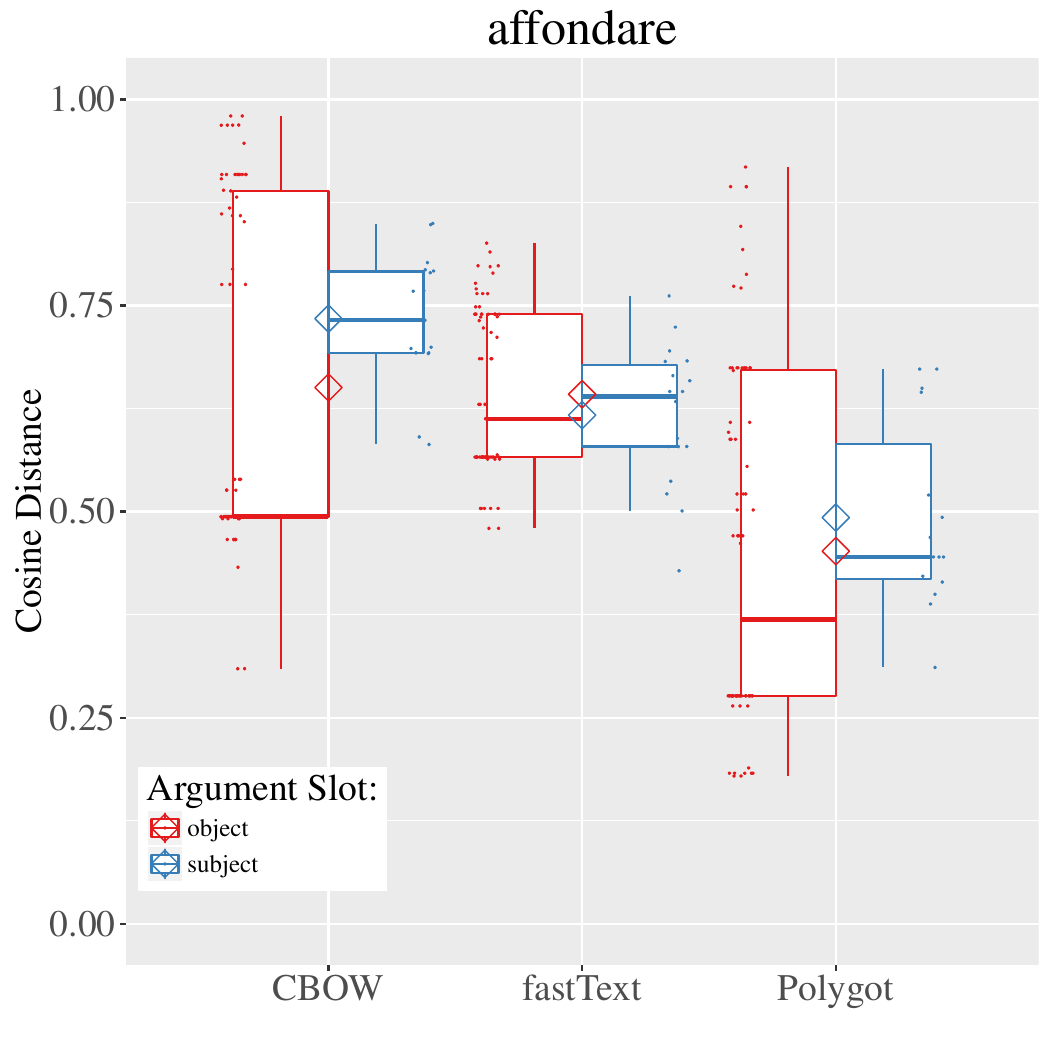}
        }%
    \end{center}
    \caption{%
        Boxes and whiskers of vector distances from centroid. Lexical sets for objects are in red, for subjects in blue. The columns in each subplot represent different models: from left to right, CBOW, fastText, and Polyglot.
     }%
\label{boxwhi}
\end{figure*}

In order to answer the questions outlined in the introduction, we devised three experiments. In all of them, we assume that the S and O lexical sets might not overlap, as borne out by \namecite{montemagni1995} and \namecite{mccarthy2000using}. The first (\S\ \ref{ssec:exp1}) investigates the internal structure of lexical sets and how their members aggregate around a prototype. Based on psycholinguistic theories (see \S\ \ref{ssec:we}), we expect vectors to lie on a continuum. As for causative-inchoative verbs in particular, objects were shown to be more homogeneous than subjects \cite{mckoon2000}, hence they should form denser clusters. The second experiment (\S\ \ref{ssec:exp2}) deals with polymorphism of lexical sets, i.e.\ the amount of distinct sub-categories they contain. We expect the number of sub-clusters to be proportional to the number of verb senses. Finally, the third experiment (\S\ \ref{ssec:exp3}) is aimed at studying how lexical sets of different arguments of the same verb are related. In particular, we predict that the distance between intransitive subject and transitive object varies depending on the spontaneity of the causative-inchoative verb (see \S\ \ref{ssec:cauinc}). In fact, intransitive subjects and transitive objects of spontaneous verbs should be limited to referents with ``teleological capability'' \cite{atkins1995building,levin1995unaccusativity}, hence creating dense groups possibly located far from each other. Fillers of agentive verbs instead should show a higher degree of overlap because of their looser constraints.

\subsection{Prototypicality: Distance from Centroid}
\label{ssec:exp1}
Once the fillers have been mapped to their corresponding vectors, a lexical set appears as a group of points in a multi-dimensional model. The centre of this group is the Euclidean mean among the vectors, which is a vector itself and is called centroid. In the first experiment, we measure the cosine distance of every vector member of a lexical set from the centroid estimated from all the other members.\footnote{Every filler was weighted proportionally to its absolute frequency.} This leave-one-out setting aims at avoiding biases due to outliers (such as phraseological usages or misspellings). In semantic terms, this measure should correspond to assessing how far a filler is from its prototype.

We obtained two sets (S and O) of cosine distance values for each verb: these can be plotted as boxes and whiskers, like in Figure \ref{boxwhi}. The cosine distances are represented as a scatter plot: on its side, a box informs about the average (horizontal line), the median\footnote{The median is the value separating the higher half of the ordered values from the lower half.} (diamond), the second and third quartiles (rectangle), and the extremes (bars) of the distribution.

Firstly, we observe a same trend for both intransitive subjects and objects that depends on the algorithms. The whole bars tend to hover around higher values for CBOW, and lower values for Polyglot. fastText instead lies somewhere in between them, and its central quartiles cover a short span of values. Secondly, there is a systematic gap between the medians for S and O. Excluding \textit{bollire}, for which data are insufficient, the object is lower 18 out of 19 times for CBOW. On the other hand, for Polyglot the preference is very mild (11 out of 18) and no clear pattern emerges for fastText.

\subsection{Polymorphism: Sub-Clusters}
\label{ssec:exp2}

\begin{figure}[b!]
     \begin{center}
        \subfigure[aumentare CBOW]{%
            \label{fig:ac}
            \includegraphics[width=0.33\textwidth]{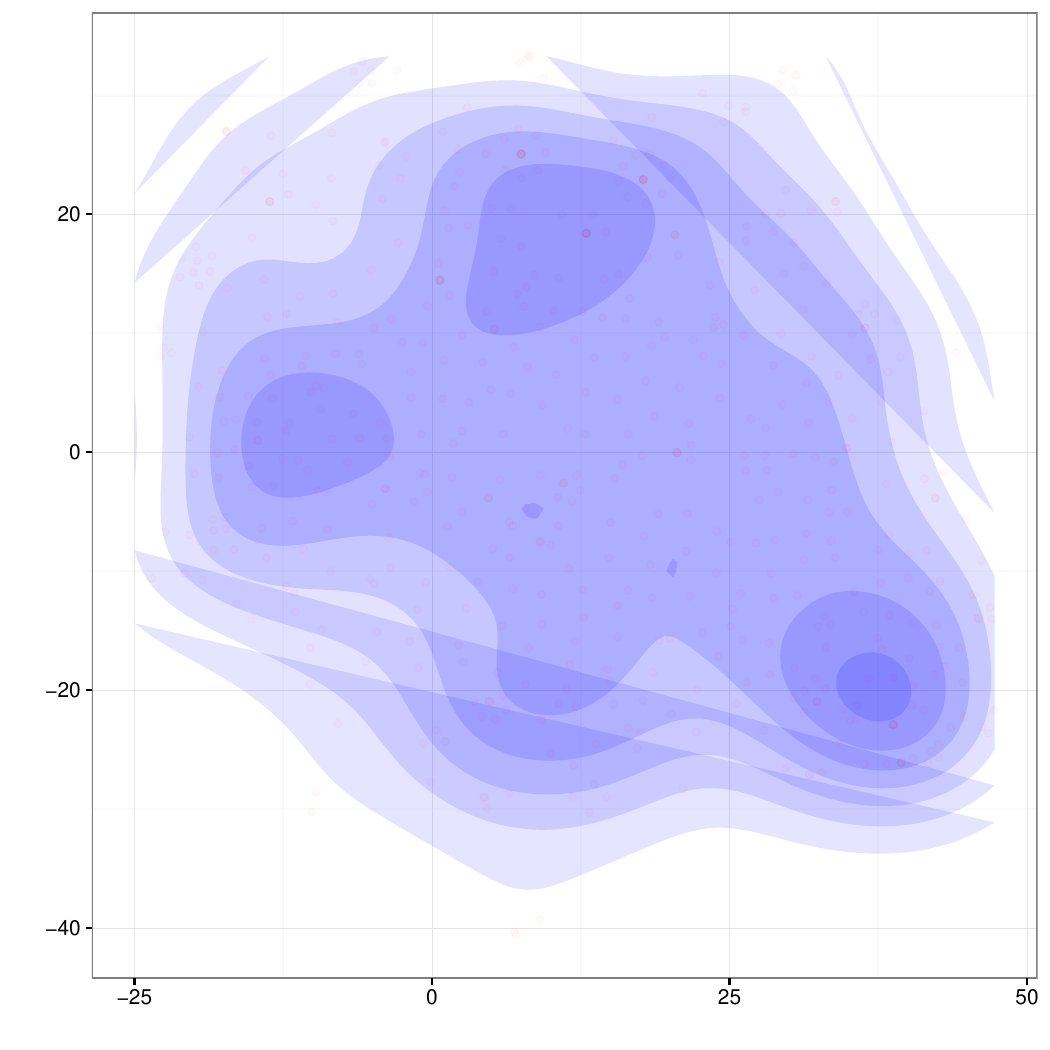}
        }%
        \subfigure[aumentare fastText]{%
           \label{fig:af}
           \includegraphics[width=0.33\textwidth]{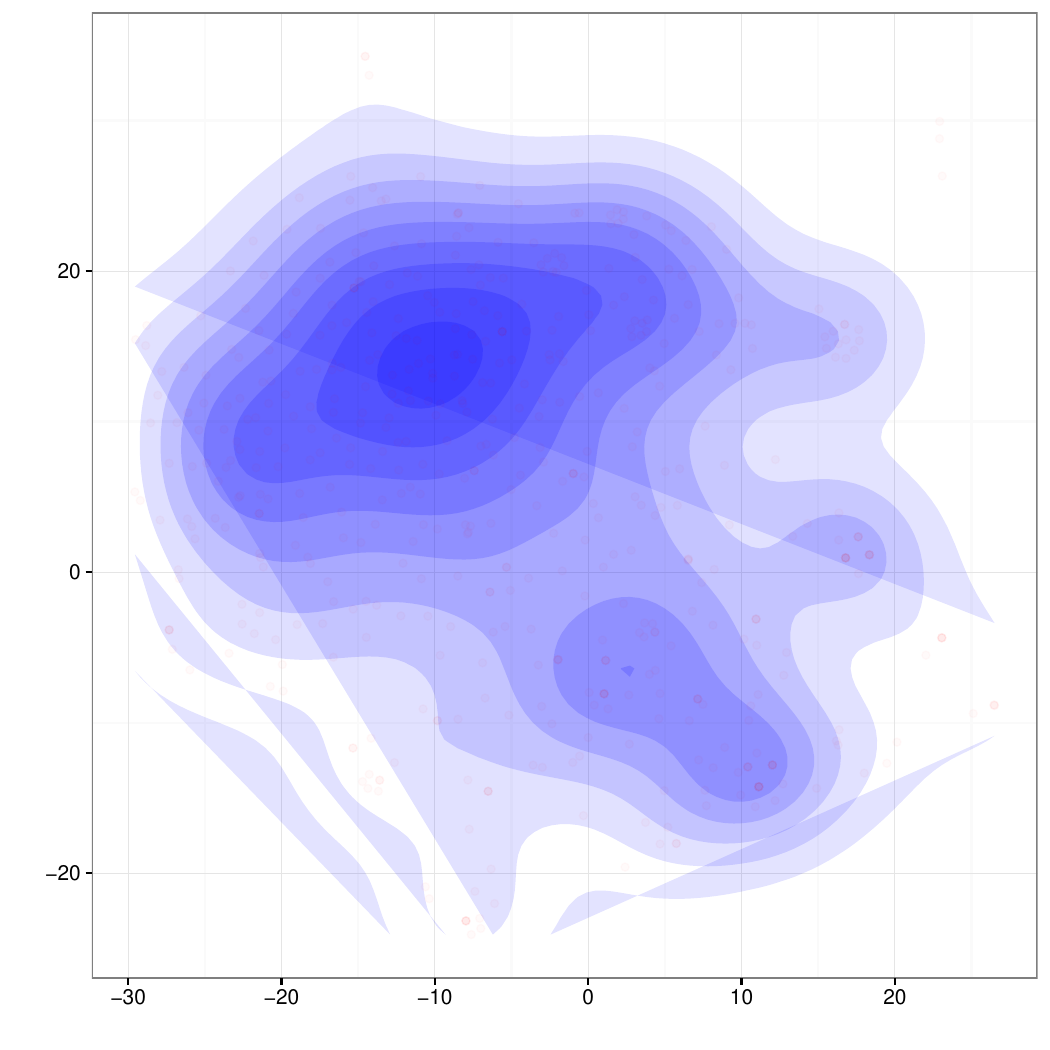}
        }%
        \subfigure[aumentare Polyglot]{%
           \label{fig:ap}
           \includegraphics[width=0.33\textwidth]{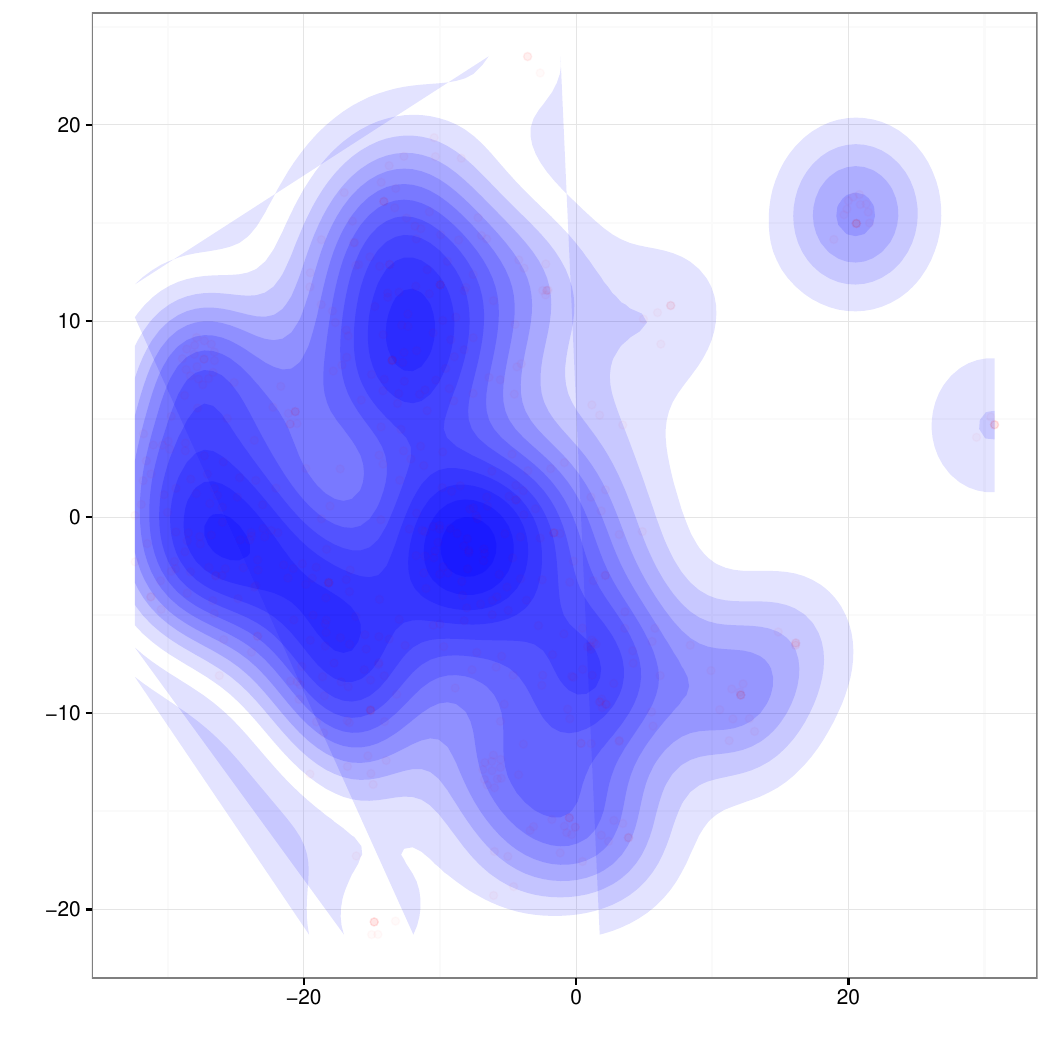}
        }%
\\%
        \subfigure[finire CBOW]{%
            \label{fig:fc}
            \includegraphics[width=0.33\textwidth]{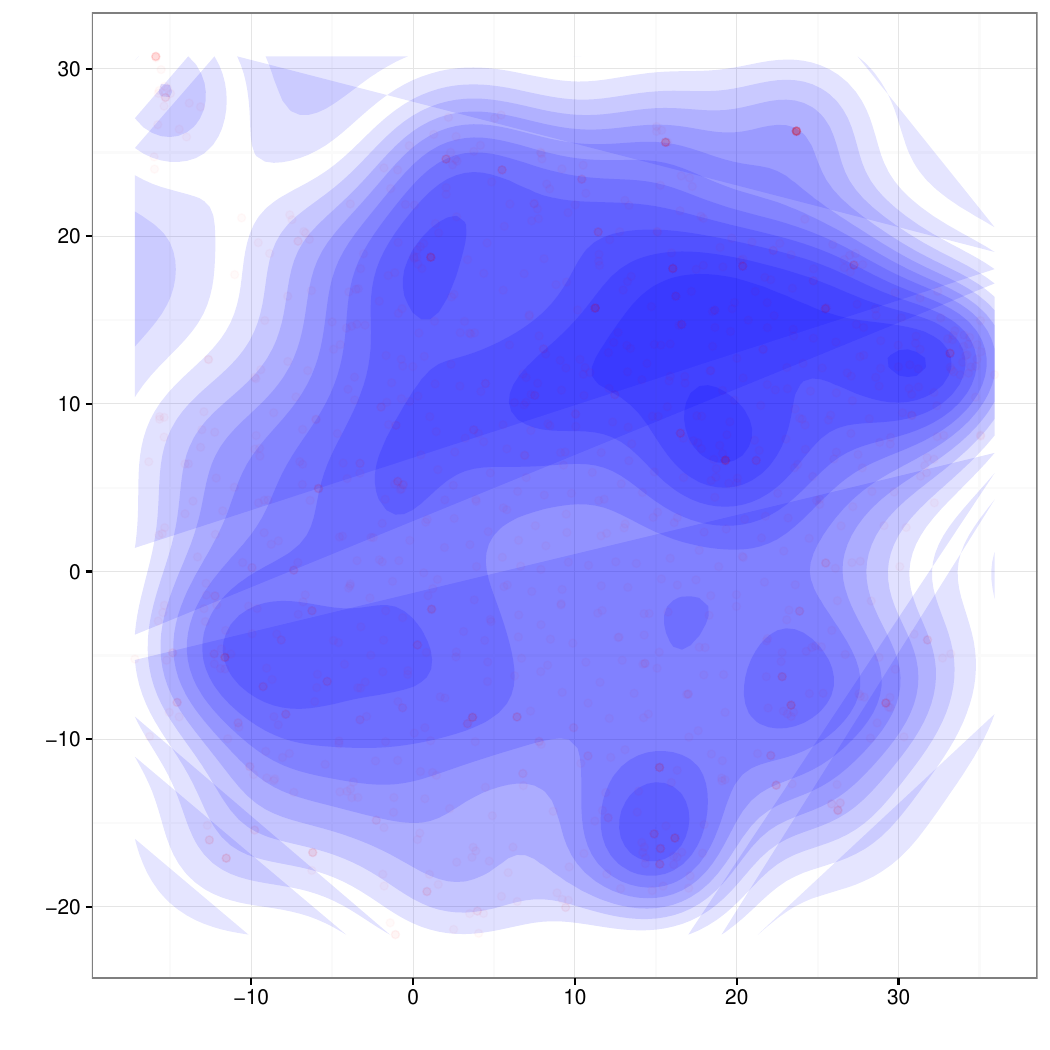}
        }%
        \subfigure[finire fastText]{%
           \label{fig:ff}
           \includegraphics[width=0.33\textwidth]{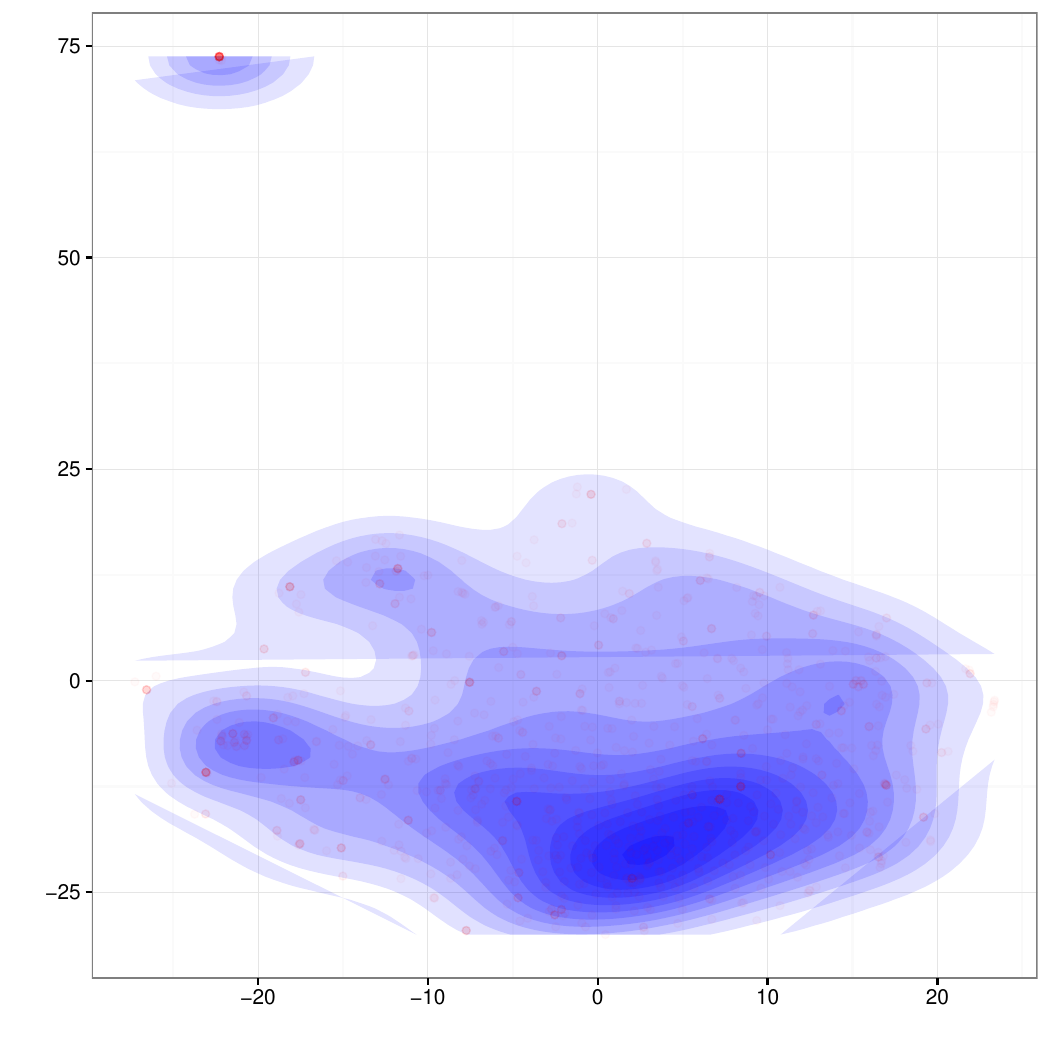}
        }%
        \subfigure[finire Polyglot]{%
           \label{fig:fp}
           \includegraphics[width=0.33\textwidth]{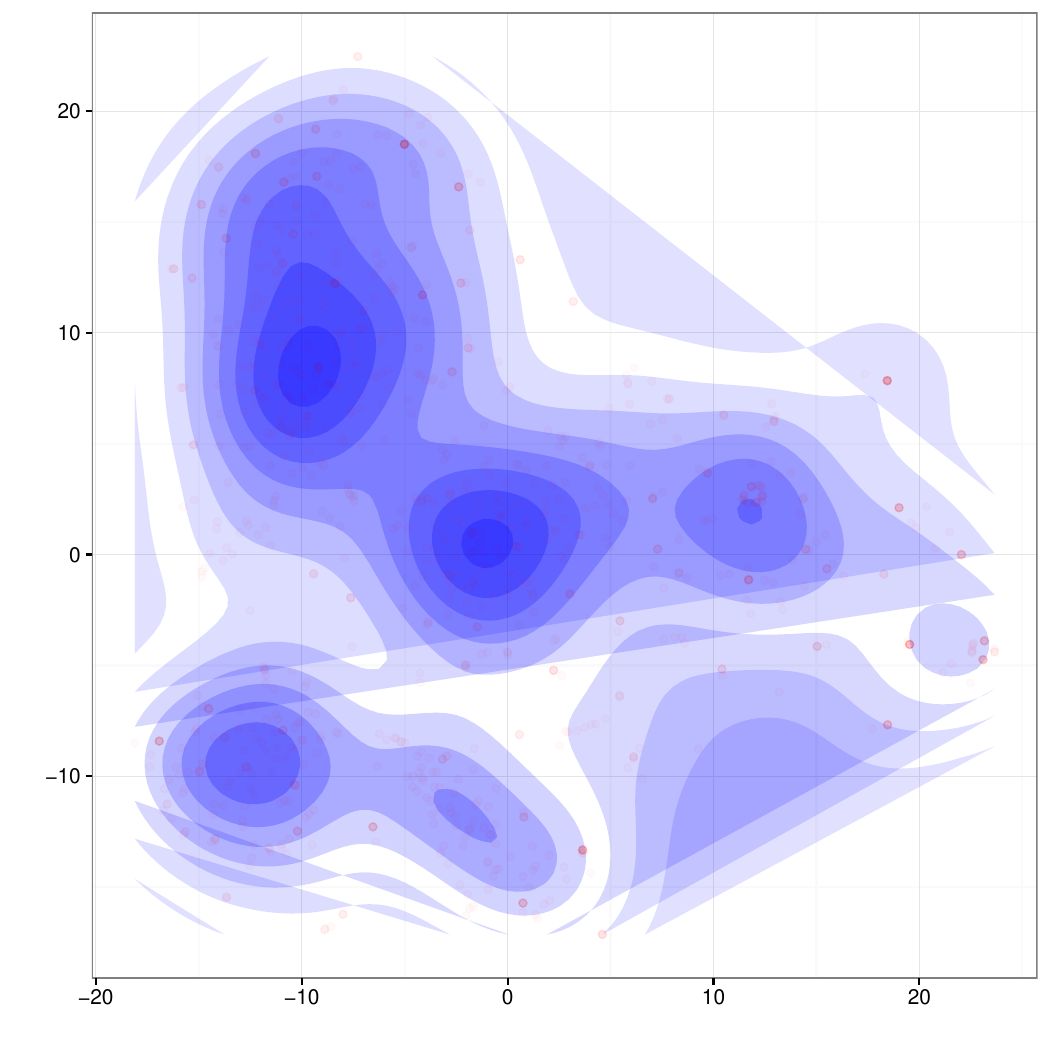}
        }%
      \end{center}
  \caption{Heatmaps of the density (low is white and high is blue) of vectors reduced to 2 dimensions through t-SNE. They belong to the S lexical sets of \textit{aumentare} (4 senses) and \textit{finire} (12 senses).}
  \label{heatmaps}
\end{figure}

The distance from the prototype examined in \S\ \ref{ssec:exp1} does not account for the complex internal structure of a lexical set, which aggregates several sub-categories. Figure \ref{heatmaps} shows a visual example of this sub-organisation by plotting the heatmaps of the density of vectors reduced to 2 dimensions through t-SNE \cite{maaten2008visualizing}: it is possible to observe spots in isolation and aggregation. In order to assess the polymorphism of lexical sets, i.e. the amount of different shades of meaning they contain, we cluster the vectors in each of them and contrast this measure with the number of the verb senses. 

Vector clustering is performed through the algorithm X-Means \cite{pelleg2000x}, an extension of the simpler algorithm K-Means \cite{macqueen1967}. The latter performs vector quantisation: it assigns vectors in a space to a pre-specified amount of \textit{k} clusters. The algorithm converges into a local optimum by initialising \textit{k} arbitrary means inside the model. Then it carries on iteratively two steps: firstly it assigns each vector to these means minimising a measure of variance. Secondly, it calculates the new means based on the newly obtained clusters. Thus, for vectors \textbf{v} clusters \textit{C} and centroids $\mu$, its objective can be formalised as:

\begin{equation}
\argmin_C \sum^{k}_{i=1} \sum_{\textbf{v} \in C_i} \| \textbf{v} - \mu_i \|
\end{equation}

In addition to these steps, X-Means can perform a further operation, namely deciding if and where splitting clusters in two to create new clusters. Iterations start from $k=2$ up to a pre-specified upper bound (in our case, $k=20$): after a split, X-Means estimates through a test whether the old cluster or the two new clusters fit the data better: in our case, this test consists in the Bayesian Information Criterion, which is defined as follows: given a split \textit{M}, the maximised value of the log-likelihood \textit{\^L} of the vectors \textit{V}, the number of parameters \textit{k}, and the number of data points \textit{n}, $ \mathrm{BIC}(M) = \hat{L}(V) - \frac{k}{2} \ln n $.

The number of clusters provided by X-Means upon convergence is displayed in Table \ref{tab:xmeans}. Moreover, we provide the number of senses of each verb according to WordNet for Italian \cite{artale1997wordnet} in a separate column. At first glance, the results show that lexical sets tend to have a similar number of clusters across the algorithms, which is surprising considering the different natures of these representations. However, this might be due to a bias: in fact, X-Means is possibly inclined to make similar decisions for sets of identical cardinality and with a low upper bound. 

We estimated the Pearson's correlation between the number of clusters and verb senses, which is reported in the bottom column: values of its coefficient mirror the strength of the correlations, ranging from -1 (negative), across 0 (absent), to 1 (positive). The p-value instead stands for the confidence by which we can exclude the null hypothesis (absence of correlation). Results for CBOW and fastText reveal a mild positive correlation. This is especially evident for subjects, which appear to be a better cue for predicting the number of verb senses. On the other hand, results are inconclusive for Polyglot, as the p-value is not significant enough.

\begin{table}[th!]
\rowcolors{1}{white}{LighterGray}
\centering
\caption{Number of verb senses and clusters inside lexical sets upon convergence of X-Means.}
\label{tab:xmeans}
\begin{tabularx}{\linewidth}{Xc cc cc cc}
\toprule
\textbf{Lemma} & \textbf{Senses} & \multicolumn{2}{c}{\textbf{CBOW}} & \multicolumn{2}{c}{\textbf{fastText}} & \multicolumn{2}{c}{\textbf{Polyglot}}\\%
\cmidrule(lr){3-4} \cmidrule(lr){5-6} \cmidrule(lr)
{7-8}

\multicolumn{2}{c}{\cellcolor{white}} & {\cellcolor{white} S} & {\cellcolor{white} O} & {\cellcolor{white} S} & {\cellcolor{white} O} & {\cellcolor{white} S} & {\cellcolor{white} O}\\
\midrule
chiudere & 9 & 10 & 13 & 15 & 13 & 10 & 13\\
aprire & 5 & 10 & 16 & 12 & 16 & 12 & 12\\
aumentare & 4 & 16 & 14 & 15 & 14 & 16 & 16\\
rompere & 6 & 9 & 14 & 10 & 14 & 10 & 13\\
riempire & 3 & 2 & 10 & 10 & 15 & 8 & 14\\
raccogliere & 6 & 2 & 15 & 8 & 14 & 2 & 14\\
connettere & 3 & 2 & 8 & 9 & 10 & 3 & 7\\
dividere & 2 & 12 & 11 & 13 & 13 & 12 & 11\\
finire & 12 & 16 & 15 & 16 & 16 & 16 & 14\\
uscire & 12 & 19 & 15 & 14 & 14 & 15 & 13\\
alzare & 6 & 6 & 12 & 10 & 16 & 11 & 13\\
scuotere & 1 & 1 & 10 & 1 & 10 & 1 & 8\\
bruciare & 11 & 8 & 13 & 13 & 14 & 2 & 11\\
congelare & 4 & 2 & 2 & 4 & 2 & 2 & 2\\
girare & 6 & 11 & 11 & 12 & 13 & 13 & 13\\
seccare & 4 & 2 & 5 & 8 & 7 & 2 & 3\\
svegliare & 1 & 7 & 8 & 9 & 7 & 10 & 8\\
sciogliere & 7 & 14 & 10 & 16 & 12 & 14 & 11\\
bollire & 2 & - & - & - & - & - & -\\
affondare & 6 & 5 & 6 & 7 & 12 & 6 & 7\\
\midrule
\multicolumn{2}{c}{\cellcolor{white} \textbf{Pearson's correlation}} & {\cellcolor{white} 0.572 } & {\cellcolor{white} 0.493 } & {\cellcolor{white} 0.596 } & {\cellcolor{white} 0.458 } & {\cellcolor{white} 0.326 } & {\cellcolor{white} 0.389 }\\
\multicolumn{2}{c}{\textbf{p-value}} & 0.011 & 0.032 & 0.007 & 0.049 & 0.173 & 0.100 \\
\end{tabularx}
\end{table} 

\begin{figure*}[th]
  \centering
  \includegraphics[width=\textwidth]{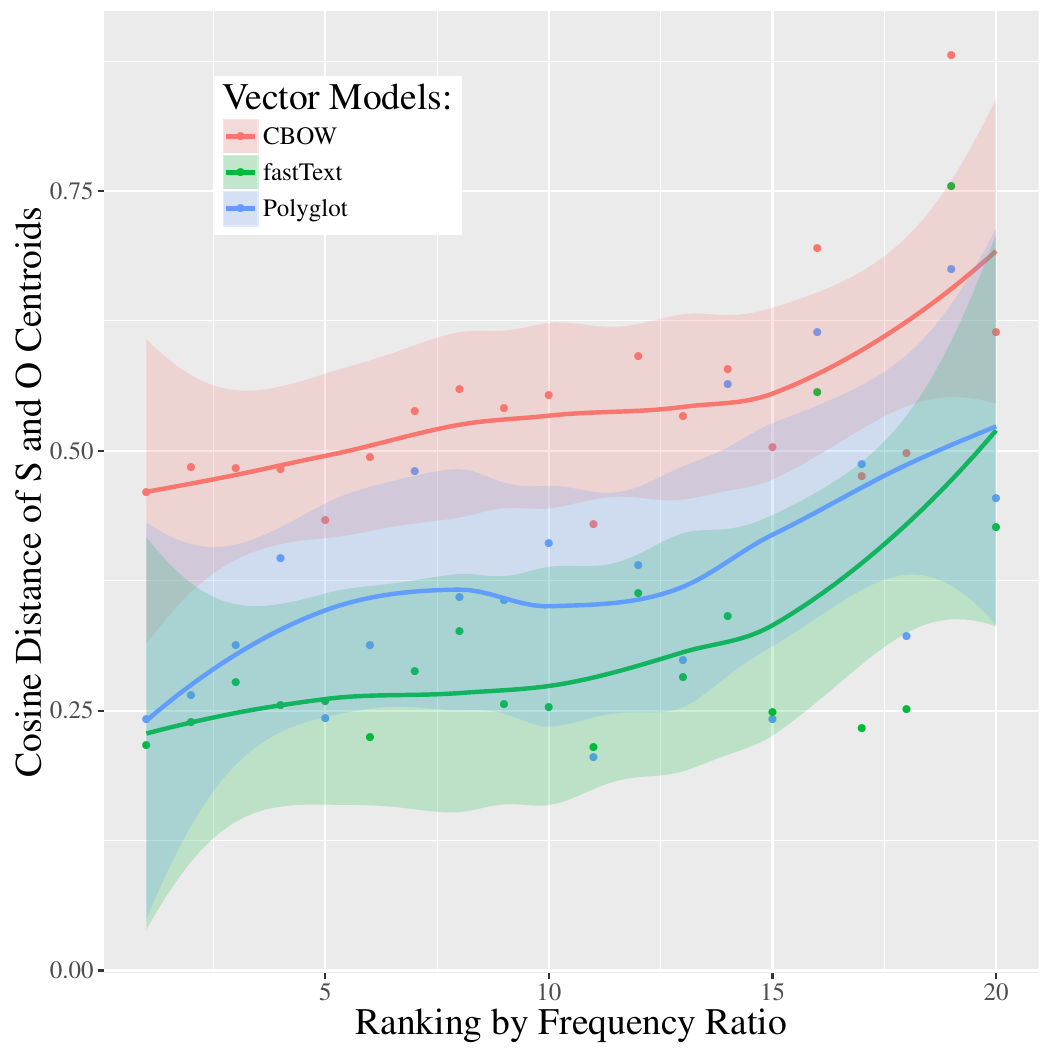}
  \caption{On the x-axis, ranking based on cross-lingual form frequencies as reported by \protect\namecite{haspelmath2014coding}. On the y-axis, cosine distances between centroid  of S and O lexical sets in Italian. Lines are LOESS regressions, and shaded areas their confidence regions.}
  \label{correlations}
\end{figure*}

\subsection{Spontaneity: Distance between Centroids}
\label{ssec:exp3}

In \S\S\ \ref{ssec:exp1}--\ref{ssec:exp2}, lexical sets of the same verb have been considered independently from each other. We now assess whether any relation holds between them by gauging the cosine distance between the centroid of S and the centroid of O for each verb. This operation is aimed at finding to which extent these two lexical sets overlap and unveiling possible asymmetries. In order to estimate whether spontaneity (see \S\ \ref{ssec:cauinc}) affects this degree of overlap, we compared the ranking of our sample of verbs according to the ratio of the cross-lingual frequency of their transitive and intransitive forms \cite{haspelmath2014coding} and a ranking based on their centroid distances.

In Figure \ref{correlations}, we plot the ranking based on cross-lingual frequency against the cosine distance: it emerges that the latter tends to increase for more spontaneous verbs, i.e. verbs with a preference for the intransitive pattern. This tendency is straightforward for all the vector models, as the LOESS line suggests. Moreover, after ranking verbs based on cosine distance between S and O lexical sets, we estimated two correlation metrics with respect to the frequency-based ranking: Spearman and Kendall. Their coefficients with the corresponding p-values are reported in Table \ref{tab:correlations}: they demonstrate a mild-strong positive correlation for both CBOW and Polyglot. However, the p-value does not allow to exclude the null hypothesis (absence of correlation) for fastText with reasonable certainty.

\begin{table}[ht]
\caption{Spearman and Kendall correlations between a ranking based on lexical set distance in Italian and another based on cross-lingual frequency ratio.}
\rowcolors{1}{white}{LighterGray}
\centering
\begin{tabularx}{0.75\linewidth}{X XXX}
\toprule &	\textbf{CBOW} &	\textbf{fastText} &	\textbf{Polyglot} \\ 
\midrule
\textbf{Spearman} & 0.560 & 0.420 & 0.490\\
\textbf{p-value} & 0.010 & 0.069 & 0.028\\
\midrule
\textbf{Kendall} & 0.411 & 0.305 & 0.358\\
\textbf{p-value} & 0.012 & 0.064 & 0.030\\
\bottomrule
\end{tabularx}
\label{tab:correlations}
\end{table}

\section{Discussion}
\label{sec:discussion}
The questions at the heart of these experiments were: how are lexical set structured? In particular, do their elements distribute uniformly in the space, or rather gather together (near or far from the prototype)? Are they polymorphic, i.e.\ composed by more than one sub-category? Moreover, which is the degree of overlap between lexical sets of different argument slots? In this section, we discuss the answers inferred from the results, and analyse the specific behaviour of every vector model.

In the first experiment (\S\ \ref{ssec:exp1}), the members of O lexical sets are scattered from the centre to the periphery. On the other hand, the members of S lexical sets lie in a more compact range of distances, mostly farther from the centroid. This implies that O behaves more similarly to a radial category, whereas S just populates the periphery. From a linguistic point of view, this means that the content of O is more homogeneous, whereas S is more heterogeneous: this finding is in line with previous work \cite{mckoon2000}. In the second experiment (\S\ \ref{ssec:exp2}), we observed a mild positive correlation between the number of clusters and of the verb senses. Hence, it is possible to conclude that polymorphic verbs accept more categories of referents as their possible argument fillers. This holds true especially for S.

In the third experiment (\S\ \ref{ssec:exp3}), we established the existence of a correlation between the verb ranking based on the cross-lingual ratio of intransitive and transitive verbs and the ranking based on cosine distance between S and O centroids in Italian. From a linguistic point of view, this is possibly due to the constraints on referents of spontaneous verbs \cite{atkins1995building,levin1995unaccusativity}, called  teleological capability \cite{copley2014theories}: this makes the sets clear-cut and possibly far from each other. This adds another piece to the puzzle of the so-called spontaneity scale: Figure \ref{fig3} shows a synopsis of our result in the context of the correlations established in previous works. Solid lines stand for correlations proven based on corpus evidence. The dotted line, on the other hand, suggests the existence of and underlying motivation for the correlations (i.e.\ spontaneity), which nonetheless remains unproven and undetermined in its nature. Its possible validation is left to future research, but remains tricky due to its purely semantic nature.

The conclusions can be drawn from more than one vector model, although results are not significant for all of them. In particular, fastText does not show any tendency with respect to the distribution of distances, nor it is possible to exclude the null hypothesis for the correlation between distance of S and O centres and frequency of the intransitive pattern. This might be due to the fact that fastText is based on characters, hence on morphological information, rather than capturing topic relatedness. Moreover, there are possible biases that plague the experiments. Firstly, the homogeneity of the O lexical set might be an artifact because the sample of objects is usually wider and hence more representative. Instead, the heterogeneity of the S lexical set is in part due to its method of extraction: sometimes also transitive subjects (A) are treated as S, because of either unexpressed objects or parsing mistakes.

\begin{figure}[t]
    \centering
\begin{tikzpicture}[->,>=stealth',shorten >=1pt,auto,node distance=3cm,
                    thick]

  \node (1) {Frequently Intransitive};
  \node (2) [above of=1] {Spontaneous};
  \node (3) [below left of=1] {Unmarked Intransitive};
  \node (4) [below right of=1] {Distant S and O centres};

  \path[->,dashed]
    (2) edge node [right] {?} (1);
  \path[<->]
    (1) edge node [right] {} (4)
    (1) edge node [left] {} (3);

\end{tikzpicture}
    \caption{Synopsis of the correlations among features of causative-inchoative verbs. The measures are based on Kendall Tau test ($\tau$) and Spearman's ranking test ($\rho$).}
    \label{fig3}
\end{figure}

\section{Conclusions and Future Work}
\label{sec:conclusion}
Our work provided evidence that lexical sets of Italian causative-inchoative verbs are non-uniform categories, whose distribution around the prototype varies to a great extent. This distribution is sensitive to the argument slot: transitive objects display a more uniform distribution of distances from the prototype, whereas the fillers of intransitive subjects lie on the edge of the category. This difference might be due to different selectional restrictions applied to the object. Moreover, the number of verb senses appeared to play a role with respect to the polymorphism of lexical sets: intuitively, the more the shades of meaning, the more the argument types that match the selectional preferences. Finally, a correlation was discovered between the cosine distance of lexical sets of a given verb in Italian and the cross-lingual behaviour of its translations, i.e.\ the tendency to appear more frequently as intransitive or as transitive. This finding has to be paired with the previously established correlation between the latter and the cross-lingual tendency to derive morphologically the intransitive form or the transitive one.

To amend the limitations mentioned in \S\ \ref{sec:discussion}
 (noise from parsing and extraction, lexical sets not fully representative), further research should: resort to an enhanced database with a wider sample, try to reduce the parsing error with state-of-art parsers, and add sense disambiguation for polysemous word forms \cite{grave2013}. Also, it should choose even more pre-trained vector models, in order to try and replicate these results. In particular, the new vector models could be optimized for similarity through semantic lexica \cite{faruqui:2015:Retro} or based on syntactic dependencies \cite{seaghdha2010latent}. The experiments in this work may be extended to other languages, either individually or through a multi-lingual word embedding model \cite{faruqui2014improving}. In fact, cross-lingually correlations are more clear-cut than those emerging single languages \cite{haspelmath2014coding}.

\section*{Acknowledgements}
This work was originally presented as a poster at \textsc{essli 2016} and then submitted to \textsc{clic-it 2016} \cite{ponti2016grounding}, where it was awarded as the `young best paper.' As a reward, it is kindly hosted by this journal in a revised and extended form. We would like to thank the anonymous reviewers for their insightful and lavish comments.

\bibliographystyle{fullname}
\bibliography{fillers}

\end{document}